\documentclass{arxiv}

\usepackage{algorithm, hyperref}
\usepackage{algpseudocode}
\usepackage{graphicx}
\usepackage{caption}
\usepackage{subcaption}
\usepackage{amsfonts}
\usepackage{url}
\usepackage{epstopdf}
\usepackage{amssymb,amsmath,subcaption}
\usepackage{longtable}
\usepackage{geometry}
\usepackage{lscape}
\usepackage{pbox}

\usepackage{color}
\usepackage{multirow}
\usepackage{multicol}

\begin{document}

\title{Benchmarking NLopt and state-of-art algorithms for Continuous Global Optimization via Hybrid IACO$_\mathbb{R}$}


\author{Udit Kumar, Sumit Soman, Jayavdeva\\
Department of Electrical Engineering, Indian Institute of Technology, Delhi, India\\
\textit{Email: udit.vlsi@gmail.com, sumit.soman@gmail.com, jayadeva@ee.iitd.ac.in}}

\begin{abstract}
This paper presents a comparative analysis of the performance of the Incremental Ant Colony algorithm for continuous optimization ($IACO_\mathbb{R}$), with different algorithms provided in the NLopt library. The key objective is to understand how the various algorithms in the NLopt library perform in combination with the Multi Trajectory Local Search (Mtsls1) technique. A hybrid approach has been introduced in the local search strategy by the use of a parameter which allows for probabilistic selection between Mtsls1 and a NLopt algorithm. In case of stagnation, the algorithm switch is made based on the algorithm being used in the previous iteration. The paper presents an exhaustive comparison on the performance of these approaches on Soft Computing (SOCO) and Congress on Evolutionary Computation (CEC) 2014 benchmarks. For both benchmarks, we conclude that the best performing algorithm is a hybrid variant of Mtsls1 with BFGS for local search.

\end{abstract}

\maketitle


\section{Introduction}

The NLopt (Non-Linear Optimization) library (v2.4.2) \cite{johnson2010nlopt} is a rich collection of optimization routines and algorithms, which provides a platform-independent interface for their use for  global and local optimization. The library has been widely used for practical implementations of optimization algorithms as well as for benchmarking new algorithms.

The work in this paper is based on the $IACO_\mathbb{R}$-LS algorithm proposed by Liao, Dorigo et al. \cite{iacor:algo} This algorithm introduced the local search procedure in the original $IACO_\mathbb{R}$ technique, specifically Mtsls1 by Tseng et al.\cite{mtsls} for local search. The $IACO_\mathbb{R}$ was an extension of the $ACO_\mathbb{R}$ algorithm for continuous optimization with the added advantage of a variable size solution archive. The premise of our work lies in improving the local search strategy adopted by $IACO_\mathbb{R}$-LS, by allowing algorithms other than Mtsls1 to be used for local search.

We present a comparison of using various algorithms from the NLopt library for local search procedure in the $IACO_\mathbb{R}$-LS algorithm. In order to introduce a hybrid approach for local search, we use a parameter that probabilistically determines whether to use the Mtsls1 algorithm or the NLopt library  algorithm. In case of stagnation, we switch between Mtsls1 or the NLopt algorithm based on the algorithm being used in the previous iteration. The objective is to rigorously analyze the effect of using various optimization algorithms in the local search procedure for $IACO_\mathbb{R}$-LS, as well as provide results on benchmark functions to enable a naive researcher to choose an algorithm easily. To the best of our knowledge, available works in literature have not provided exhaustive comparisons using optimization algorithm libraries on ant colony based approaches, other than \cite{rios2013derivative}. However,  surveys on state-of-art in multi-objective evolutionary algorithms \cite{zhou2011multiobjective}, differential evolution \cite{das2011differential} and real-parameter evolutionary multimodal optimization \cite{das2011real} have appeared in literature.

The rest of the paper is organized as follows. Section \ref{sec:hybrid} discusses our hybrid approach which allows using Mtsls1 alongwith an NLopt library algorithm for local search phase of $IACO_\mathbb{R}$-LS. This is followed by a discussion on the NLopt library in Section \ref{sec:nlopt}. We present our results and a discussion in Section \ref{sec:results}, followed by the conclusions in Section \ref{sec:conclusion}. 

\section{Hybrid Local Search using Mtsls1 and NLopt algorithms }
\label{sec:hybrid}

We begin by introducing the Mtsls1 algorithm, and the motivation to develop a hybrid approach for local search. This is followed by a description of our algorithm which uses the hybrid local search using Mtsls1 and the algorithms from the NLopt library.

The Multi-Trajectory Local Search, or Mtsls1 algorithm \cite{mtsls} exploits the search space across multiple paths. The approach has evolved to many variants, notable among which are the self-adaptive evolution by Zhao et al. \cite{zhao2011self}, multi-objective optimization \cite{tseng2009multiple} and dynamic search trajectories by Snyman et al. \cite{snyman1987multi}.

Mtsls1 searches along one dimension, and optimum value of one dimension is used as starting point for the next dimension. At each dimension, Mtsls1 tries to move by a step size $s$ along one dimension, and evaluates the change in the function value. If the function value decreases, then new point is used for optimization along the next dimension, If the function value increases, then algorithm goes back to the starting point and moves by a factor of the step size, $0.5*s$ towards negative direction and evaluates the function. Again the function value is compared and based on minimum value of the function, the optimum point is provided. 

We propose an hybrid local-search approach which incorporates the non-gradient based Mtsls1, alongwith an algorithm from the NLopt library as part of the $IACO_\mathbb{R}-LS$ technique. Our approach offers a choice between selecting either of the two, based on a probabilistically determined choice. Ths is indicated by the algorithm parameter $P(nlopt)$. In case this probabilistic choice fails to provide any improvement after a specific number of iterations, we switch the algorithm being used based on the algorithm used in the previous iteration. The parameters $ctr_{localsearch}$ and $thresh_{localsearch}$ have been used in our algorithm implement this, as we select a different local search algorithm when $ctr_{localsearch}$ crosses $thresh_{localsearch}$. This ensures that our local-search approach does not stagnate, and also gives our approach an ``adaptive'' flavor.

All algorithms from NLopt library are used as part of the hybrid local search approach. The Nlopt algorithms meant for global optimization are allowed as many function evaluations as set for global search, but for local search, maximum allowed function evaluations in a single local search call is set to $160$. It may be noted here that the method for approximating the gradient is directly linked to algorithm's ability to escape local minima. Solomon \cite{salomon1998evolutionary} opines that \textit{``if, however, the gradient is estimated by independent trials with a distance along each axis, the difference between both classes of algorithms almost vanishes.''} Hence, our computations of gradient are based on approximating the derivative using central differences. By using this method, the maximum number of function evaluations would effectively be ($2*n$ \textit{(for gradient)}$+1$ \textit{(function evaluation)})$*160$) for local search, for an $n$-dimensional problem.  Our approach is illustrated in Algorithm \ref{ouralgo}; note that hybrid local-search approach is incorporated at Steps (\ref{ourlocalsearch})-(\ref{ourlocalsearch2}), for additional details reader may refer to \cite{mtsls}.

\begin{algorithm}[H]
\caption{Hybrid $IACO_\mathbb{R}$ Algorithm}\label{ouralgo}
\begin{algorithmic}[1]
\Procedure{Hybrid $IACO_\mathbb{R}$}{Probability ($p$), constant parameter ($\zeta$), initial archive size ($\alpha$), growth ($\gamma$), maximum archive size ($\alpha_{max}$), function tolerance ($\tau$), 
maximum failure ($Fail_{max}$), maximum stagnation iterations ($Siter_{max}$), dimensions ($N$), termination criteria ($Tc$), probability of switching to NLopt algorithm ($P(nlopt)$), maximum local search iterations ($thresh_{localsearch}$)}
\State \text{Initialize }$\alpha$ \text{ solutions}
\State \text{Evaluate initial solutions}
%
%
\While{($Tc$ not satisfied)}
 \If {($Fail_{(i,best)} < Fail_{max}$)}
		\State \text{Local search from }$Sol_{best}$

\ElsIf{($Fail_{(i,random)} < Fail_{max}$)}
		\State \text{Local search from }$Sol_{random}$
		\EndIf
 \If {$P(nlopt) == 0$}\Comment{Local Search}\label{ourlocalsearch}
		\State \text{Use Mtsls1}

\ElsIf{$P(nlopt) == 1$}
		\State \text{Use NLopt algo}
		\Else
\If{$ctr_{localsearch} < thresh_{localsearch}$}
			 			\If{$rand()<P(nlopt)$}
								\State \text{Use NLopt algo}
						\Else
								\State \text{Use Mtsls1}
						\EndIf
			\Else
						\If{\text{Last iteration used Mtsls1}}
								\State \text{Use NLopt algo}
						\Else
								\State \text{Use Mtsls1}
						
\EndIf
			\EndIf
\EndIf \label{ourlocalsearch2}

 \If{\text{No improvement in solution}}
		\State \text{Increment }$Fail_i$
	\EndIf

\Comment{Continued...}
\algstore{myalg}
\end{algorithmic}
\end{algorithm}

\begin{algorithm}[H]
\begin{algorithmic}[1]
\algrestore{myalg}

\If{$rand() < p$}\Comment{Generate new solution}
		\State \text{Sample best Gaussian for each dimension}

 \If{New soln is better}\Comment{Exploitation}
				\State \text{Substitute new solution for }$Sol_{best}$
		 \EndIf
\Else
		\ForAll{$j \in [1:\alpha]$}\Comment{Exploration}
		\State \text{Sample Gaussian along each dimension }$j$ 
		\If{new solution is better}
				\State \text{Substitute new solution for }$Sol_{j}$
		 \EndIf
		\EndFor
\EndIf

\If{$Iter_{curr}$ is a multiple of $\gamma$ and $\alpha < \alpha_{max}$}\Comment{Archive Growth}
		\State \text{Initialize new solutions using }$S_{new} = S_{new} + rand() \cdot(S_{best}-S_{new})$
		\State \text{Add new solution to archive}
		\State \text{Increment }$\alpha$
\EndIf

\If{$ctr_{global search}$==$Siter_{max}$}\Comment{Restart}
		\State \text{Re-initialize solution set without } $S_{best}$
\EndIf
\EndWhile
\EndProcedure

\end{algorithmic}
\end{algorithm}

\section{The NLopt Library }
\label{sec:nlopt}
    
\begin{figure}[h]
\centering
\includegraphics[scale=0.8]{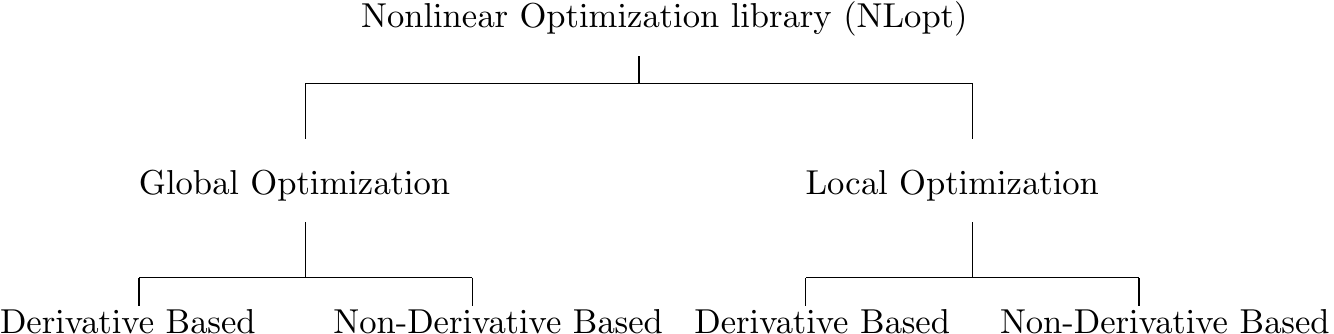}
\caption{Categories of algorithms in the NLopt library}
\label{nlopt_libs}
\end{figure}

The NLopt library optimization algorithms are partitioned into four categories as shown in Figure \ref{nlopt_libs}; algorithms in each category are listed in Table \ref{nlopt_algos}. For the sake of brevity, each algorithm has been assigned a numeric identifier in Table \ref{nlopt_algos} (Col. \textit{``ID''}) which is used to refer to them in the subsequent sections of this paper.

\begin{table*}[htbp]
  \centering  
    \caption{NLopt algorithms}
    \scalebox{0.7}{
    \begin{tabular}{|c|c|l|l|}
    \hline
    \multicolumn{4}{|c|}{\textbf{Summary of Nlopt Algorithms}} \\
    \hline
    \multicolumn{1}{|c|}{S. No. } & \multicolumn{1}{|c|}{ID} & \multicolumn{1}{|c|}{Algorithm} & \multicolumn{1}{|c|}{Code} \\
    \hline
    \hline
    \multicolumn{4}{|c|}{\textit{\textbf{Global Search Algorithms (Non Derivative Based)}}} \\
    \hline
    1     & A0    & DIRECT & NLOPT\_GN\_DIRECT \\
    2     & A1    & DIRECT-L & NLOPT\_GN\_DIRECT\_L \\
    3     & A2    & Randomized DIRECT-L & NLOPT\_GN\_DIRECT\_L\_RAND \\
    4     & A3    & Unscaled DIRECT & NLOPT\_GN\_DIRECT\_NOSCAL \\
    5     & A4    & Unscaled DIRECT-L & NLOPT\_GN\_DIRECT\_L\_NOSCAL \\
    6     & A5    & Unscaled Randomized DIRECT-L & NLOPT\_GN\_DIRECT\_L\_RAND\_NOSCAL \\
    7     & A6    & Original DIRECT version & NLOPT\_GN\_ORIG\_DIRECT \\
    8     & A7    & Original DIRECT-L version & NLOPT\_GN\_ORIG\_DIRECT\_L \\
    9     & A19   & Controlled random search (CRS2) with local mutation & NLOPT\_GN\_CRS2\_LM \\
    10    & A20   & Multi-level single-linkage (MLSL), random & NLOPT\_GN\_MLSL \\
    11    & A22   & Multi-level single-linkage (MLSL), quasi-random & NLOPT\_GN\_MLSL\_LDS \\
    12    & A35   & ISRES evolutionary constrained optimization & NLOPT\_GN\_ISRES \\
    13    & A36   & Augmented Lagrangian method & NLOPT\_AUGLAG \\
    14    & A37   & Augmented Lagrangian method for equality constraints & NLOPT\_AUGLAG\_EQ \\
    15    & A38   & Multi-level single-linkage (MLSL), random & NLOPT\_G\_MLSL \\
    16    & A39   & Multi-level single-linkage (MLSL), quasi-random & NLOPT\_G\_MLSL\_LDS \\
    17    & A42   & ESCH evolutionary strategy & NLOPT\_GN\_ESCH \\
    \hline
    \multicolumn{4}{|c|}{\textit{\textbf{Global Search Algorithms (Derivative Based)}}} \\
    \hline
    18    & A8   & Stochastic Global Optimization (StoGO) & NLOPT\_GD\_STOGO \\
    19    & A9   & Stochastic Global Optimization (StoGO), random & NLOPT\_GD\_STOGO\_RAND \\
    20    & A21   & Multi-level single-linkage (MLSL), random & NLOPT\_GD\_MLSL \\
    21    & A23   & Multi-level single-linkage (MLSL) quasi-random & NLOPT\_GD\_MLSL\_LDS \\
    \hline
    \multicolumn{4}{|c|}{\textit{\textbf{Local Search Algorithms (Non Derivative Based)}}} \\
    \hline
    22    & A12   & Principal-axis, PRAXIS & NLOPT\_LN\_PRAXIS \\
    23    & A25   & COBYLA & NLOPT\_LN\_COBYLA \\
    24    & A26   & NEWUOA unconstrained optimization via quadratic models & NLOPT\_LN\_NEWUOA \\
    25    & \textit{A27*}   & \textit{Bound-constrained optimization via NEWUOA-based quadratic models} & \textit{NLOPT\_LN\_NEWUOA\_BOUND} \\
    26    & A28   & Nelder-Mead simplex algorithm & NLOPT\_LN\_NELDERMEAD \\
    27    & A29   & Sbplx variant of Nelder-Mead & NLOPT\_LN\_SBPLX \\
    28    & A30   & Augmented Lagrangian method & NLOPT\_LN\_AUGLAG \\
    29    & A32   & Augmented Lagrangian method for equality constraints & NLOPT\_LN\_AUGLAG\_EQ \\
    30    & A34   & BOBYQA bound-constrained optimization via quadratic models & NLOPT\_LN\_BOBYQA \\
    \hline
    \multicolumn{4}{|c|}{\textit{\textbf{Local Search Algorithms (Derivative Based)}}} \\
    \hline
     31	  & \textit{A10**} & \textit{Original L-BFGS by Nocedal et al.}	& \textit{NLOPT\_LD\_LBFGS\_NOCEDAL}\\
    32   & A11   & Limited-memory BFGS & NLOPT\_LD\_LBFGS \\
    33    & A13   & Limited-memory variable-metric, rank 1 & NLOPT\_LD\_VAR1 \\
    34    & A14   & Limited-memory variable-metric, rank 2 & NLOPT\_LD\_VAR2 \\
    35    & A15   & Truncated Newton & NLOPT\_LD\_TNEWTON \\
    36    & A16   & Truncated Newton with restarting & NLOPT\_LD\_TNEWTON\_RESTART \\
    37    & A17   & Preconditioned truncated Newton & NLOPT\_LD\_TNEWTON\_PRECOND \\
    38    & A18   & Preconditioned truncated Newton with restarting & NLOPT\_LD\_TNEWTON\_PRECOND\_RESTART \\
    39    & A24   & Method of Moving Asymptotes (MMA) & NLOPT\_LD\_MMA \\
    40    & A31   & Augmented Lagrangian method & NLOPT\_LD\_AUGLAG \\
    41    & A33   & Augmented Lagrangian method for equality constraints & NLOPT\_LD\_AUGLAG\_EQ \\
    42    & A40   & Sequential Quadratic Programming (SQP) & NLOPT\_LD\_SLSQP \\
    43    & A41   & CCSA with simple quadratic approximations & NLOPT\_LD\_CCSAQ \\
    \hline
      \multicolumn{4}{|c|}{\textit{* - This algorithm has not been considered in this study as runtime errors were encountered during execution.}}\\
      \multicolumn{4}{|c|}{\textit{** - The original algorithm is not part of NLopt library, after minor modification it has been made part of $A11$.}}\\

    \hline
    \end{tabular}%
    }
  \label{nlopt_algos}%
\end{table*}%

The global search algorithms can be categorized into derivative and non-derivative based algorithms. All global optimization algorithms require bound constraints to be specified on the optimization parameters \cite{mtsls}. The DIviding RECTangles (DIRECT) algorithm proposed by Jones et al. \cite{jones1993lipschitzian, finkel2003direct} is based on dividing the search space into hyperrectangles and searching simultaneously at the global and local level. We consider the DIRECT (A0), unscaled DIRECT (A3) and original DIRECT (A6) versions of the algorithm. A locally biased variant of the DIRECT approach proposed by  Gablonsky et al. \cite{gablonsky2001locally} is the DIRECT-L, which is expected to perform well on functions with a single global minima and few local minima. We consider the DIRECT-L (A1), randomized DIRECT-L (A2), unscaled DIRECT-L (A4), unscaled randomized DIRECT-L (A5) and the original DIRECT-L (A7) versions of this algorithm in our study.

The Controlled Random Search (CRS) algorithm with local mutation by Kaelo et al. \cite{kaelo2006some} is similar to the idea behind genetic algorithms, where an initial population evolves across generations to converge to the minima. This algorithm uses an evolution strategy similar the Nelder Mead algorithm \cite{lagarias1998convergence}. The version of the CRS algorithm provided in the NLopt library supports bound constraints and starts with an initial population size of $10*(n+1)$ for an $n$ dimensional problem.  We use the CRS2 with local mutation (A19) for our study.

The Multi-Level Single Linkage (MLSL) algorithm by Kan and Timmer \cite{kan1987stochastic} is based on selecting multiple start points initially at random, and then using clustering heuristics to traverse the search space effeciently without redundancy. The algorithm configuration in the NLopt library allows for sampling four random points by default, with default function and variable tolerances set to $10^{-15}$ and $10^{-7}$ respectively. We have used the non-derivative-based random MLSL (A20) and quasi-random MLSL (A22) for global search, their derivative-independent versions (A38 and A39), and derivative-based versions (A21 and A23) respectively.

The Improved Stochastic Ranking Evolution Strategy (ISRES) algorithm by Runarsson and Yao \cite{runarsson2005search} supports optimization with both linear and nonlinear constraints (A35). The algorithm uses a mutation rule with log-normal step size and an update rule similar to the Nelder Mead method. The default configuration for the initial population size in the NLopt library is $20*(n+1)$ for an $n$-dimensional function. Another evolutionary algorithm available in the library and used in our study is the ESCH algorithm by Santos et al. \cite{santos2010designing} which supports only linear bound constraints (A42). The STOchastic Global Optimization algorithm (StoGO) by Madsen et al. \cite{madsen1998global, zertchaninov1998c, gudmundsson1998parallel}  is a derivative-based global search algorithm which supports only bound constraints. We consider the original StoGO (A8) and its randomized variant (A9).

In the category of non-gradient based local search methods, the Constrained Optimization BY Linear Approximation (COBYLA) by Powell \cite{powell1994direct} supports non-linear equality and inequality constraints (A25). It is based on linearly approximating the objective function using a simplex of $(n+1)$ points for an $n$-dimensional problem. Another version of this algorithm which supports bound constraints is Bound Optimization BY Quadratic Approximation, referred to as BOBYQA \cite{powell2009bobyqa} (A34). Its enhanced version is NEWUOA \cite{powell2006newuoa, powell2008developments, powell2007developments} (A26), with support for constrained and unconstrained problems. The PRincipal AXIS algorithm (PRAXIS) by Brent \cite{brent1973algorithms} primarily supports unconstrained optimization (A12). Other algorithms available in this category in the NLopt library that have been used in this study include the well known Nelder-Mead Simplex method \cite{nelder1965simplex} (A28), and its subplex variant by Rowan \cite{rowan1990functional} (A29).

We now present a brief summary of the derivative based algorithms for local search. The Broyden-Fletcher-Goldfarb-Shanno (BFGS) \cite{wright1999numerical} algorithm (A11) has been a classical optimization algorithm belonging to the class of approximate Newton methods. Along with several variants \cite{wei2004superlinear, wan2014new, liu1989limited, li2001modified}, BFGS has been widely used in the domain of unconstrained optimization. The Method of Moving Asymptotes (MMA) algorithm (A24) by Svanberg \cite{svanberg2002class} is based on locally approximating the gradient of the objective function and a quadratic penalty term. It is an enhanced version of the original Conservative Complex Separable Approximation (CCSA) algorithm \cite{svanberg2002class} (A41), which provides for pre-conditioning of the Hessian in the version available as part of the NLopt library. The Sequential Quadratic Programming (SQP) by Kraft \cite{kraft1988software, kraft1994algorithm} (A40) is available for both linear and non-linear equality and inequality constraints. The preconditioned truncated Newton method \cite{dembo1983truncated} allows for using gradient information from previous iterations, which provides for faster convergence with the trade-off for greater memory requirement. We have used the truncated Newton method (A15), with restart (A16), pre-conditioned Newton method (A17) and pre-conditioned with restart (A18) for our study.

A limited memory variable metric algorithm by Vl{\v{c}}ek and Luk{\v{s}}an \cite{vlvcek2006shifted} is available with rank-1 (A13) and rank-2 (A14) methods in the NLopt library. Also, an Augmented Lagrangian algorithm by Conn et al. \cite{conn1991globally, birgin2008improving} is available for all categories including gradient/non-gradient and global/local search. This algorithm combines the objective and associated constraints into a single function with a penalty term. This is solved separately as another problem without non-linear constraints, to finally converge to the desired solution. Variants of this algorithm to consider penalty function for only equality constraints is also available in the library. We have used the Augmented Lagrangian method (A30), a version with equality constraint support (A32), and the corresponding derivative based versions A31 and A33 respectively. We now present the results of our comparative analysis using these algorithms alongwith Mtsls1 for local search in the following section.

\section{Study and Discussion }
\label{sec:results}

We evaluate the performance of our approach by comparing its performance with the algorithms featured in SOCO and CEC 2014 benchmarks. The results are categorized into three subsections, presenting the results on the SOCO, CEC 2014 benchmarks, and a comparison of the standalone performance of the NLopt algorithms with our hybrid approach. For our study, we use $P(nlopt)$ as 0.6, and parameters for the NLopt algorithms as $x_{tol_{rel}} = x_{tol_{abs}} = 1e-7$, $f_{tol_{rel}} = f_{tol_{abs}} = 1e-15$ (for description of these parameters, the reader may refer to \cite{johnson2010nlopt}). Parameters specific to algorithms have been used with their default configuration, while ranking parameters for performance have been elucidated in \cite{chenproblem},

\subsection{Results on SOCO benchmarks}
We used the 50-dimensional versions of the 19 benchmark functions suite shown in Table \ref{soco_benchmark}. Functions $F1-F6$ were originally proposed for the special session on large scale global optimization organized for the IEEE 2008 Congress on Evolutionary Computation (CEC 2008) \cite{socof1-f6}, Functions $F7-F11$ were proposed at the ISDA 2009 Conference. Functions $F12-F19$ are hybrid functions that combine two functions belonging to $F1-F11$.  Some properties of the benchmark functions are listed in Table \ref{soco_benchmark}. The detailed description of these functions is available in \cite{socof1-f19, socof1-press}.

\begin{table}
\caption{SOCO Benchmark functions}
\label{soco_benchmark}
\centering
\scalebox{0.7}{
\begin{tabular}{|c|c|c|c|c|c|c|c|} 
\hline ID & Name & Analytical Form & \pbox{20cm}{Uni(U)/\\Multi(M)\\Modal} & Sep. & Rotated & \pbox{20cm}{Easily \\ optimized \\ dimension-\\ wise}\\
\hline $F1$ & Shift Sphere & $\sum_{i=1}^D z_i^2 + f_{bias}, z=x-o$ & U & Y & N & Y\\
\hline $F2$ & Shifted Schwefel 2.21 & $max_i \lbrace |z_i|, 1 \leq i \leq D \rbrace + f_{bias}, z=x-o$ & U & N & N & N \\
\hline $F3$ & Shifted Rosenbrock & $\sum_{i=1}^{D-1} (100 (z_i^2+z_{i+1})^2+(z_i-1)^2))+ f_{bias}, z=x-o$ & M & N & N &  Y\\
\hline $F4$ & Shift. Rastrigin &$\sum_{i=1}^D (z_i^2 - 10 cos(2 \pi z_i)+10)+ f_{bias}, z=x-o$  & M & Y & N &  Y\\
\hline $F5$ & Shift. Griewank & $\sum_{i=1}^D \frac{z_i^2}{4000} - \prod_{i=1}^D cos(\frac{z_i}{\sqrt{i}})+1+f_{bias}, z=x-o$ & M & N & N &  N\\
\hline $F6$ & Shift. Ackley & \pbox{20cm}{$-20 e^{-0.2 \sqrt{\frac{1}{D}\sum_{i=1}^Dz_i^2}} -e^{\frac{1}{D}\sum_{i=1}^Dcos(2 \pi z_i)}+$ \\ $20+e+f_{bias}, z=x-o$}  & M & Y & N &  Y\\
\hline $F7$ & Shift. Schwefel 2.22 & $\sum_{i=1}^D |z_i| + \prod_{i=1}^D |z_i|, z=x-o$  & U & Y & N  & Y\\
\hline $F8$ & Shift Schwefel 1.2 & $\sum_{i=1}^D (\sum_{j=1}^i z_j)^2, z=x-o$ & U & N & N &  N\\
\hline $F9$ & Shift. Extended $F_{10}$  & \pbox{20cm}{$\sum_{i=1}^{D-1}f_{10}(z_i,z_{i+1})+f_{10}(z_D,z_1), z=x-o$\\where $f_{10}=(x^2+y^2)^{0.25} (sin^2(50(x^2+y^2)^{0.1})+1)$}& U & N & N  & Y\\
\hline $F10$ & Shift. Bohachevsky  & \pbox{20cm}{$\sum_{i=1}^D z_i^2 + 2z_{i+1}^2-0.3cos(3\pi z_i)-0.4cos(4\pi z_{i+1})+$ \\ $0.7, z=x-o$ }& U & N & N  & N\\
\hline $F11$ & Shift. Schafler  & $\sum_{i=1}^{D-1} (z_i^2+z_{i+1}^2)^{0.25} (sin^2(50(z_i^2+z_{i+1}^2)^{0.1})+1), z=x-o$ & U & N & N  & Y\\
\hline $F12$ & Hybrid Function & $F9$ + 0.25 $F1$ & M & N & N  & N\\
\hline $F13$ & Hybrid Function & $F9$ + 0.25 $F3$ &  M & N & N  & N\\
\hline $F14$ & Hybrid Function & $F9$ + 0.25 $F4$ &  M & N & N  & N\\
\hline $F15$ & Hybrid Function  & $F10$ + 0.25 $F7$ &  M & N & N  & N\\
\hline $F16$ & Hybrid Function  & $F9$ + 0.5 $F1$ &  M & N & N  & N\\
\hline $F17$ & Hybrid Function  & $F9$ + 0.75 $F3$ &  M & N & N  & N\\
\hline $F18$ & Hybrid Function  & $F9$ + 0.75 $F4$ &  M & N & N  & N\\
\hline $F19$ & Hybrid Function  & $F10$ + 0.75 $F7$ &  M & N & N  & N\\
\hline \end{tabular}
}
\end{table}

We applied the termination conditions used for SOCO, that is, the maximum number of function evaluations was $5000\times D$, where $D$ denotes the number of dimensions in which the function is considered. All the investigated algorithms were run 25 times on each function. We report error values defined as $f(x)-f(x^*)$, where $x$ is a candidate solution and $x^*$ is the optimal solution. Error values lower than $10^{-14}$ (this value is referred to as 0-threshold) are approximated to 0. Our analysis is based on either the whole solution quality distribution, or on the median and average errors. For the evaluation of our $IACO_\mathbb{R}$-Hybrid approach, we use the algorithm parameters as indicated in Table 2 of \cite{iacor:algo}.

The average and median errors obtained on the benchmark functions have been shown in Tables \ref{nlopt_avg_tab} and \ref{nlopt_med_tab} respectively. The algorithms from the NLopt library (used for hybridization within $IACO_\mathbb{R}$-Mtsls1 framework) have been indicated as the rows of the table using numeric identifiers provided in Table \ref{nlopt_algos}, while the SOCO benchmark functions are provided as the columns. These provide a comprehensive analysis of the performance of these algorithms on the functions; cases where the error is zero have been indicated in boldface. It may be noted here that only the  Differential Evolution algorithm (DE) \cite{storn1997differential}, the co-variance matrix adaptation evolution strategy with increasing population size (G-CMA-ES) \cite{auger2005restart} and the real-coded CHC algorithm (CHC) \cite{eshelman1993chapter} have been considered as baseline algorithms for performance evaluation on SOCO benchmarks in \cite{socof1-f19, socof1-press}. Further, the source code for implementation of IACO$_\mathbb{R}$ is available online at http://iridia.ulb.ac.be/supp/IridiaSupp2011-008/.

\begin{landscape}
\begin{table*}
\caption{Average error for NLopt algorithms for 50D}
\label{nlopt_avg_tab}
\scalebox{0.5}{
\begin{tabular} {|c|c|c|c|c|c|c|c|c|c|c|c|c|c|c|c|c|c|c|c|}
\hline ALGOs & F1 & F2 & F3 & F4 & F5 & F6 & F7 & F8 & F9 & F10 & F11 & F12 & F13 & F14 & F15 & F16 & F17 & F18 & F19 \\
\hline A0 & 2.04E-13 & 6.23E+01 & 2.15E+02 & 2.64E+01 & 2.33E-14 & 1.43E+00 & 3.00E-02 & 1.95E+01 & 6.69E-01 & 1.51E-01 & 5.69E-02 & 1.17E-03 & 3.69E+02 & 7.07E+01 & 9.34E-01 & 1.08E-02 & 5.58E+03 & 3.91E+01 & \textbf{0.00E+00} \\
\hline A1 & 5.94E-14 & 2.51E-12 & 1.61E+02 & 7.44E+01 & 5.62E-03 & 8.90E-02 & 2.09E-13 & 1.76E+03 & 7.76E+00 & 1.89E-01 & 4.47E+01 & 6.05E-14 & 1.47E+02 & 5.23E+01 & 1.49E+00 & 4.54E-14 & 3.96E+03 & 5.92E+00 & 2.96E+00 \\
\hline A2 & \textbf{0.00E+00} & 5.11E+01 & 1.38E+02 & 4.26E+00 & 5.42E-03 & 1.41E-01 & \textbf{0.00E+00} & 1.80E-01 & 1.13E-01 & \textbf{0.00E+00} & 5.64E-02 & \textbf{0.00E+00} & 7.32E+00 & 2.26E+00 & \textbf{0.00E+00} & \textbf{0.00E+00} & 1.02E+03 & 4.93E-01 & \textbf{0.00E+00} \\
\hline A3 & 5.94E-14 & 1.28E-12 & 1.61E+02 & 7.39E+01 & 5.62E-03 & 8.90E-02 & 2.20E-13 & 1.76E+03 & 7.76E+00 & 1.89E-01 & 4.47E+01 & 6.05E-14 & 1.47E+02 & 5.43E+01 & 1.49E+00 & 4.54E-14 & 3.96E+03 & 6.15E+00 & 2.96E+00 \\
\hline A4 & 5.94E-14 & 1.28E-12 & 1.61E+02 & 7.39E+01 & 5.62E-03 & 8.90E-02 & 2.20E-13 & 1.76E+03 & 7.76E+00 & 1.89E-01 & 4.47E+01 & 6.05E-14 & 1.47E+02 & 5.43E+01 & 1.49E+00 & 4.54E-14 & 3.96E+03 & 6.15E+00 & 2.96E+00 \\
\hline A5 & 5.94E-14 & 1.28E-12 & 1.61E+02 & 7.39E+01 & 5.62E-03 & 8.90E-02 & 2.20E-13 & 1.76E+03 & 7.76E+00 & 1.89E-01 & 4.47E+01 & 6.05E-14 & 1.47E+02 & 5.43E+01 & 1.49E+00 & 4.54E-14 & 3.96E+03 & 6.15E+00 & 2.96E+00 \\
\hline A6 & 4.96E+00 & 3.69E+01 & 1.10E+04 & 1.14E+02 & 6.81E-01 & 8.90E-02 & 2.42E+00 & 5.51E+03 & 1.26E+02 & 4.69E-01 & 1.32E+02 & 1.30E+02 & 3.68E+04 & 7.88E+01 & 7.07E+00 & 8.53E+01 & 4.65E+03 & 3.91E+01 & 5.61E+00 \\
\hline A7 & \textbf{0.00E+00} & 8.27E-13 & 2.11E+02 & 1.13E+02 & 5.62E-03 & 8.90E-02 & 2.23E-14 & 5.51E+03 & 1.26E+02 & 2.98E-01 & 1.32E+02 & \textbf{0.00E+00} & 1.94E+02 & 7.87E+01 & 3.17E+00 & \textbf{0.00E+00} & 4.58E+03 & 3.91E+01 & 5.53E+00 \\
\hline A8 & \textbf{0.00E+00} & 3.57E-12 & 6.34E+01 & 7.56E-01 & \textbf{0.00E+00} & 6.89E-14 & \textbf{0.00E+00} & 2.48E-05 & \textbf{0.00E+00} & \textbf{0.00E+00} & 1.05E-02 & 4.25E-02 & 3.05E+00 & 5.50E-01 & \textbf{0.00E+00} & 5.36E-03 & 1.30E+02 & 4.64E-02 & \textbf{0.00E+00} \\
\hline A9 & \textbf{0.00E+00} & 3.57E-12 & 6.34E+01 & 7.56E-01 & \textbf{0.00E+00} & 6.89E-14 & \textbf{0.00E+00} & 2.48E-05 & \textbf{0.00E+00} & \textbf{0.00E+00} & 1.05E-02 & 4.25E-02 & 3.05E+00 & 5.50E-01 & \textbf{0.00E+00} & 5.36E-03 & 1.30E+02 & 4.64E-02 & \textbf{0.00E+00} \\
\hline A11 & \textbf{0.00E+00} & \textbf{0.00E+00} & 1.13E-11 & \textbf{0.00E+00} & \textbf{0.00E+00} & \textbf{0.00E+00} & \textbf{0.00E+00} & \textbf{0.00E+00} & \textbf{0.00E+00} & \textbf{0.00E+00} & \textbf{0.00E+00} & 7.27E-03 & 1.86E+00 & 4.37E-01 & \textbf{0.00E+00} & \textbf{0.00E+00} & 6.89E+00 & 1.63E-01 & \textbf{0.00E+00} \\
\hline A12 & \textbf{0.00E+00} & 1.53E-14 & 8.20E+01 & 3.98E-02 & \textbf{0.00E+00} & 1.14E-14 & \textbf{0.00E+00} & 3.07E-09 & \textbf{0.00E+00} & \textbf{0.00E+00} & 3.75E-02 & \textbf{0.00E+00} & 3.38E+00 & 3.90E-01 & \textbf{0.00E+00} & 6.64E-03 & 1.70E+02 & 2.62E-03 & \textbf{0.00E+00} \\
\hline A13 & \textbf{0.00E+00} & \textbf{0.00E+00} & 8.11E-14 & 1.75E+00 & \textbf{0.00E+00} & \textbf{0.00E+00} & \textbf{0.00E+00} & \textbf{0.00E+00} & \textbf{0.00E+00} & \textbf{0.00E+00} & 7.82E-02 & 5.15E-03 & 1.49E+00 & 1.12E+00 & \textbf{0.00E+00} & 4.29E-04 & 8.01E+01 & 1.02E+00 & \textbf{0.00E+00} \\
\hline A14 & \textbf{0.00E+00} & \textbf{0.00E+00} & 8.11E-14 & 1.75E+00 & \textbf{0.00E+00} & \textbf{0.00E+00} & \textbf{0.00E+00} & \textbf{0.00E+00} & \textbf{0.00E+00} & \textbf{0.00E+00} & 7.82E-02 & 5.15E-03 & 1.49E+00 & 1.12E+00 & \textbf{0.00E+00} & 4.29E-04 & 8.01E+01 & 1.02E+00 & \textbf{0.00E+00} \\
\hline A15 & \textbf{0.00E+00} & \textbf{0.00E+00} & 3.19E-01 & 5.97E-01 & \textbf{0.00E+00} & \textbf{0.00E+00} & \textbf{0.00E+00} & \textbf{0.00E+00} & 2.33E-01 & \textbf{0.00E+00} & 7.84E-01 & 1.71E-02 & 6.37E+00 & 2.85E+00 & \textbf{0.00E+00} & 1.51E-01 & 8.49E+01 & 2.57E+00 & \textbf{0.00E+00} \\
\hline A16 & \textbf{0.00E+00} & \textbf{0.00E+00} & 9.57E-01 & 1.68E-13 & \textbf{0.00E+00} & \textbf{0.00E+00} & \textbf{0.00E+00} & \textbf{0.00E+00} & 4.88E-02 & \textbf{0.00E+00} & 6.32E-04 & 6.00E-02 & 6.81E+00 & 1.69E+00 & \textbf{0.00E+00} & 1.84E-01 & 4.10E+01 & 1.30E+00 & \textbf{0.00E+00} \\
\hline A17 & \textbf{0.00E+00} & \textbf{0.00E+00} & 4.78E-01 & 7.96E-02 & 3.95E-04 & \textbf{0.00E+00} & \textbf{0.00E+00} & \textbf{0.00E+00} & 3.97E+00 & \textbf{0.00E+00} & 1.96E+00 & 2.60E-02 & 8.10E+00 & 1.75E+00 & \textbf{0.00E+00} & 2.00E+00 & 3.78E+01 & 3.10E+00 & \textbf{0.00E+00} \\
\hline A18 & \textbf{0.00E+00} & \textbf{0.00E+00} & 7.97E-01 & 1.59E+00 & 6.57E-13 & \textbf{0.00E+00} & \textbf{0.00E+00} & \textbf{0.00E+00} & 4.35E-01 & \textbf{0.00E+00} & 6.95E-01 & 6.98E-02 & 2.47E+00 & 1.42E+00 & \textbf{0.00E+00} & 1.68E-02 & 6.57E+01 & 3.52E-01 & \textbf{0.00E+00} \\
\hline A19 & 2.34E-13 & 1.72E+01 & 1.53E+02 & 5.66E+01 & 8.77E-03 & 1.37E-01 & 1.83E-06 & 2.95E+01 & 1.03E+02 & 2.24E+00 & 1.06E+02 & 4.71E+01 & 1.06E+02 & 6.97E+01 & 3.37E-01 & 1.35E+02 & 2.25E+02 & 4.67E+01 & 2.21E+00 \\
\hline A20 & 9.60E-11 & 2.56E+01 & 1.73E+02 & 9.99E+01 & 6.87E-10 & 1.80E+00 & 9.59E+00 & 4.58E-06 & 2.42E+02 & 1.09E+01 & 2.55E+02 & 5.88E+01 & 8.80E+01 & 1.06E+02 & 1.39E+01 & 1.30E+02 & 9.07E+02 & 7.73E+01 & 1.39E+01 \\
\hline A21 & \textbf{0.00E+00} & 3.75E+01 & 2.07E+02 & 3.98E+02 & \textbf{0.00E+00} & 1.10E+01 & 2.25E+07 & 3.06E+01 & 2.76E+02 & 1.18E+01 & 2.80E+02 & 7.64E+01 & 1.03E+02 & 3.22E+02 & 9.38E+02 & 1.66E+02 & 4.23E+02 & 1.48E+02 & 1.31E+01 \\
\hline A22 & 1.07E-10 & 2.61E+01 & 1.52E+02 & 1.04E+02 & 6.82E-10 & 1.78E+00 & 5.94E+00 & 4.90E-06 & 2.47E+02 & 1.08E+01 & 2.50E+02 & 5.97E+01 & 8.86E+01 & 8.88E+01 & 1.47E+01 & 1.30E+02 & 1.30E+03 & 7.11E+01 & 1.39E+01 \\
\hline A23 & \textbf{0.00E+00} & 3.70E-03 & 1.07E+02 & 3.03E+00 & \textbf{0.00E+00} & \textbf{0.00E+00} & \textbf{0.00E+00} & 5.52E-01 & 2.37E+00 & \textbf{0.00E+00} & 1.93E+00 & 4.39E-01 & 1.90E+01 & 3.32E+00 & 2.62E-13 & 1.58E+00 & 2.06E+01 & 2.69E+00 & \textbf{0.00E+00} \\
\hline A24 & \textbf{0.00E+00} & 6.37E-03 & 1.81E+02 & \textbf{0.00E+00} & \textbf{0.00E+00} & \textbf{0.00E+00} & \textbf{0.00E+00} & 5.60E-01 & \textbf{0.00E+00} & \textbf{0.00E+00} & 5.92E-02 & 1.54E-01 & 9.38E+00 & 1.92E+00 & \textbf{0.00E+00} & 1.08E-01 & 2.03E+02 & 2.39E-01 & \textbf{0.00E+00} \\
\hline A25 & \textbf{0.00E+00} & 3.68E-12 & 7.69E+01 & 4.78E-01 & \textbf{0.00E+00} & 6.91E-14 & \textbf{0.00E+00} & 3.34E-05 & \textbf{0.00E+00} & \textbf{0.00E+00} & 5.87E-02 & 7.04E-02 & 2.10E+00 & 9.85E-03 & \textbf{0.00E+00} & \textbf{0.00E+00} & 1.50E+02 & \textbf{0.00E+00} & \textbf{0.00E+00} \\
\hline A26 & \textbf{0.00E+00} & 3.31E-12 & 7.07E+01 & 7.96E-02 & \textbf{0.00E+00} & 6.82E-14 & \textbf{0.00E+00} & 3.37E-05 & 3.47E-02 & \textbf{0.00E+00} & 1.05E-02 & 8.72E-03 & 2.74E+00 & 2.88E-01 & \textbf{0.00E+00} & \textbf{0.00E+00} & 2.31E+01 & \textbf{0.00E+00} & \textbf{0.00E+00} \\
\hline A28 & \textbf{0.00E+00} & 3.40E-12 & 1.35E+02 & 6.77E-01 & \textbf{0.00E+00} & 7.25E-14 & \textbf{0.00E+00} & 5.09E-05 & 1.18E-01 & \textbf{0.00E+00} & 3.47E-02 & 9.32E-02 & 2.71E+00 & 3.66E-01 & \textbf{0.00E+00} & 1.82E-01 & 1.80E+02 & 1.59E-01 & \textbf{0.00E+00} \\
\hline A29 & \textbf{0.00E+00} & 3.54E-12 & 4.35E+01 & 6.37E-01 & \textbf{0.00E+00} & 6.79E-14 & \textbf{0.00E+00} & 2.93E-05 & 2.08E-03 & \textbf{0.00E+00} & 4.64E-02 & 7.51E-02 & 3.25E+00 & 4.33E-01 & \textbf{0.00E+00} & \textbf{0.00E+00} & 8.55E+01 & 3.18E-01 & \textbf{0.00E+00} \\
\hline A30 & \textbf{0.00E+00} & 3.68E-12 & 7.69E+01 & 4.78E-01 & \textbf{0.00E+00} & 6.91E-14 & \textbf{0.00E+00} & 3.35E-05 & \textbf{0.00E+00} & \textbf{0.00E+00} & 5.87E-02 & 7.04E-02 & 2.10E+00 & 9.85E-03 & \textbf{0.00E+00} & \textbf{0.00E+00} & 1.50E+02 & \textbf{0.00E+00} & \textbf{0.00E+00} \\
\hline A31 & \textbf{0.00E+00} & 6.37E-03 & 1.81E+02 & \textbf{0.00E+00} & \textbf{0.00E+00} & \textbf{0.00E+00} & \textbf{0.00E+00} & 5.60E-01 & \textbf{0.00E+00} & \textbf{0.00E+00} & 5.92E-02 & 1.54E-01 & 9.38E+00 & 1.92E+00 & \textbf{0.00E+00} & 1.08E-01 & 2.03E+02 & 2.39E-01 & \textbf{0.00E+00} \\
\hline A32 & \textbf{0.00E+00} & 3.68E-12 & 7.69E+01 & 4.78E-01 & \textbf{0.00E+00} & 6.91E-14 & \textbf{0.00E+00} & 3.35E-05 & \textbf{0.00E+00} & \textbf{0.00E+00} & 5.87E-02 & 7.04E-02 & 2.10E+00 & 9.85E-03 & \textbf{0.00E+00} & \textbf{0.00E+00} & 1.50E+02 & \textbf{0.00E+00} & \textbf{0.00E+00} \\
\hline A33 & \textbf{0.00E+00} & 6.37E-03 & 1.81E+02 & \textbf{0.00E+00} & \textbf{0.00E+00} & \textbf{0.00E+00} & \textbf{0.00E+00} & 5.60E-01 & \textbf{0.00E+00} & \textbf{0.00E+00} & 5.92E-02 & 1.54E-01 & 9.38E+00 & 1.92E+00 & \textbf{0.00E+00} & 1.08E-01 & 2.03E+02 & 2.39E-01 & \textbf{0.00E+00} \\
\hline A34 & \textbf{0.00E+00} & 3.57E-12 & 1.01E+02 & 7.56E-01 & \textbf{0.00E+00} & 6.89E-14 & \textbf{0.00E+00} & 2.94E-05 & \textbf{0.00E+00} & \textbf{0.00E+00} & 1.05E-02 & 2.60E-03 & 3.06E+00 & 5.50E-01 & \textbf{0.00E+00} & 5.36E-03 & 1.81E+02 & 4.64E-02 & \textbf{0.00E+00} \\
\hline A35 & 6.74E+02 & 1.17E+01 & 8.15E+06 & 2.44E+02 & 7.24E+00 & 4.01E+00 & 1.61E+01 & 6.14E+03 & 1.59E+02 & 7.39E+01 & 1.60E+02 & 2.92E+02 & 1.54E+06 & 1.67E+02 & 2.77E+01 & 2.10E+02 & 5.89E+02 & 7.42E+01 & 3.97E+01 \\
\hline A36 & \textbf{0.00E+00} & 3.57E-12 & 6.34E+01 & 7.56E-01 & \textbf{0.00E+00} & 6.89E-14 & \textbf{0.00E+00} & 2.48E-05 & \textbf{0.00E+00} & \textbf{0.00E+00} & 1.05E-02 & 4.25E-02 & 3.05E+00 & 5.50E-01 & \textbf{0.00E+00} & 5.36E-03 & 1.30E+02 & 4.64E-02 & \textbf{0.00E+00} \\
\hline A37 & \textbf{0.00E+00} & 3.57E-12 & 6.34E+01 & 7.56E-01 & \textbf{0.00E+00} & 6.89E-14 & \textbf{0.00E+00} & 2.48E-05 & \textbf{0.00E+00} & \textbf{0.00E+00} & 1.05E-02 & 4.25E-02 & 3.05E+00 & 5.50E-01 & \textbf{0.00E+00} & 5.36E-03 & 1.30E+02 & 4.64E-02 & \textbf{0.00E+00} \\
\hline A38 & \textbf{0.00E+00} & 3.57E-12 & 6.34E+01 & 7.56E-01 & \textbf{0.00E+00} & 6.89E-14 & \textbf{0.00E+00} & 2.48E-05 & \textbf{0.00E+00} & \textbf{0.00E+00} & 1.05E-02 & 4.25E-02 & 3.05E+00 & 5.50E-01 & \textbf{0.00E+00} & 5.36E-03 & 1.30E+02 & 4.64E-02 & \textbf{0.00E+00} \\
\hline A39 & \textbf{0.00E+00} & 3.57E-12 & 6.34E+01 & 7.56E-01 & \textbf{0.00E+00} & 6.89E-14 & \textbf{0.00E+00} & 2.48E-05 & \textbf{0.00E+00} & \textbf{0.00E+00} & 1.05E-02 & 4.25E-02 & 3.05E+00 & 5.50E-01 & \textbf{0.00E+00} & 5.36E-03 & 1.30E+02 & 4.64E-02 & \textbf{0.00E+00} \\
\hline A40 & \textbf{0.00E+00} & \textbf{0.00E+00} & 4.33E+00 & 9.66E-01 & \textbf{0.00E+00} & \textbf{0.00E+00} & \textbf{0.00E+00} & \textbf{0.00E+00} & 1.18E-01 & \textbf{0.00E+00} & 9.78E-01 & \textbf{0.00E+00} & 2.35E+00 & 1.24E+00 & \textbf{0.00E+00} & \textbf{0.00E+00} & 1.90E+02 & \textbf{0.00E+00} & \textbf{0.00E+00} \\
\hline A41 & \textbf{0.00E+00} & 9.23E-03 & 4.88E+01 & 3.98E-02 & \textbf{0.00E+00} & \textbf{0.00E+00} & \textbf{0.00E+00} & 7.20E-01 & 5.63E+00 & \textbf{0.00E+00} & 2.10E-02 & \textbf{0.00E+00} & 1.71E+01 & 1.54E-01 & \textbf{0.00E+00} & 5.57E-01 & 2.60E+02 & 8.07E-02 & \textbf{0.00E+00} \\
\hline A42 & \textbf{0.00E+00} & 4.48E-12 & 6.94E+01 & 1.59E-01 & \textbf{0.00E+00} & 6.92E-14 & \textbf{0.00E+00} & 3.40E-05 & \textbf{0.00E+00} & \textbf{0.00E+00} & 5.69E-02 & 1.29E-02 & 2.49E+00 & 4.80E-02 & \textbf{0.00E+00} & 5.36E-03 & 1.19E+02 & 1.21E-01 & \textbf{0.00E+00} \\
\hline 
\end{tabular}
}
\end{table*}
\end{landscape}

\begin{landscape}
\begin{table*}
\caption{Median error for NLopt algorithms for 50D}
\label{nlopt_med_tab}
\scalebox{0.5}{
\begin{tabular} {|c|c|c|c|c|c|c|c|c|c|c|c|c|c|c|c|c|c|c|c|}
\hline ALGOs & F1 & F2 & F3 & F4 & F5 & F6 & F7 & F8 & F9 & F10 & F11 & F12 & F13 & F14 & F15 & F16 & F17 & F18 & F19 \\
\hline A0 & \textbf{0.00E+00} & 9.66E+01 & 8.19E+01 & 1.49E+01 & \textbf{0.00E+00} & 2.56E+00 & 3.61E-11 & 9.73E+00 & 9.34E-05 & \textbf{0.00E+00} & 1.49E-08 & 9.22E-04 & 5.32E+02 & 8.87E+01 & 1.41E-12 & 1.07E-02 & 7.61E+03 & 4.38E+01 & \textbf{0.00E+00} \\
\hline A1 & \textbf{0.00E+00} & 2.40E-12 & 1.11E+02 & 8.06E+01 & 7.40E-03 & 1.59E-01 & 3.74E-13 & 5.90E+02 & 1.26E+00 & 1.90E-12 & 2.67E+01 & 1.08E-13 & 1.53E+02 & 5.55E+01 & 2.18E-09 & 8.11E-14 & 5.35E+03 & 3.98E+00 & 1.16E-07 \\
\hline A2 & \textbf{0.00E+00} & 7.66E+01 & 3.10E+01 & \textbf{0.00E+00} & \textbf{0.00E+00} & 7.51E-14 & \textbf{0.00E+00} & 8.91E-03 & \textbf{0.00E+00} & \textbf{0.00E+00} & \textbf{0.00E+00} & \textbf{0.00E+00} & 3.77E+00 & \textbf{0.00E+00} & \textbf{0.00E+00} & \textbf{0.00E+00} & 8.59E+01 & \textbf{0.00E+00} & \textbf{0.00E+00} \\
\hline A3 & \textbf{0.00E+00} & 1.22E-12 & 1.11E+02 & 8.06E+01 & 7.40E-03 & 1.59E-01 & 3.92E-13 & 5.90E+02 & 1.26E+00 & 1.90E-12 & 2.67E+01 & 1.08E-13 & 1.53E+02 & 5.55E+01 & 2.18E-09 & 8.11E-14 & 5.35E+03 & 3.98E+00 & 1.16E-07 \\
\hline A4 & \textbf{0.00E+00} & 1.22E-12 & 1.11E+02 & 8.06E+01 & 7.40E-03 & 1.59E-01 & 3.92E-13 & 5.90E+02 & 1.26E+00 & 1.90E-12 & 2.67E+01 & 1.08E-13 & 1.53E+02 & 5.55E+01 & 2.18E-09 & 8.11E-14 & 5.35E+03 & 3.98E+00 & 1.16E-07 \\
\hline A5 & \textbf{0.00E+00} & 1.22E-12 & 1.11E+02 & 8.06E+01 & 7.40E-03 & 1.59E-01 & 3.92E-13 & 5.90E+02 & 1.26E+00 & 1.90E-12 & 2.67E+01 & 1.08E-13 & 1.53E+02 & 5.55E+01 & 2.18E-09 & 8.11E-14 & 5.35E+03 & 3.98E+00 & 1.16E-07 \\
\hline A6 & 8.26E+00 & 5.68E+01 & 1.91E+04 & 1.26E+02 & 1.06E+00 & 1.59E-01 & 4.04E+00 & 5.99E+03 & 1.62E+02 & 7.54E-01 & 1.63E+02 & 2.02E+02 & 6.13E+04 & 1.06E+02 & 1.04E+01 & 1.20E+02 & 6.31E+03 & 4.38E+01 & 9.32E+00 \\
\hline A7 & \textbf{0.00E+00} & 8.10E-13 & 2.55E+02 & 1.25E+02 & 7.40E-03 & 1.59E-01 & 2.32E-14 & 5.99E+03 & 1.62E+02 & 4.70E-01 & 1.63E+02 & \textbf{0.00E+00} & 2.59E+02 & 1.06E+02 & 4.67E+00 & \textbf{0.00E+00} & 6.22E+03 & 4.38E+01 & 9.19E+00 \\
\hline A8 & \textbf{0.00E+00} & 1.75E-12 & 1.88E+01 & \textbf{0.00E+00} & \textbf{0.00E+00} & 7.51E-14 & \textbf{0.00E+00} & 2.16E-05 & \textbf{0.00E+00} & \textbf{0.00E+00} & \textbf{0.00E+00} & \textbf{0.00E+00} & 2.06E+00 & \textbf{0.00E+00} & \textbf{0.00E+00} & \textbf{0.00E+00} & 4.48E+01 & \textbf{0.00E+00} & \textbf{0.00E+00} \\
\hline A9 & \textbf{0.00E+00} & 1.75E-12 & 1.88E+01 & \textbf{0.00E+00} & \textbf{0.00E+00} & 7.51E-14 & \textbf{0.00E+00} & 2.16E-05 & \textbf{0.00E+00} & \textbf{0.00E+00} & \textbf{0.00E+00} & \textbf{0.00E+00} & 2.06E+00 & \textbf{0.00E+00} & \textbf{0.00E+00} & \textbf{0.00E+00} & 4.48E+01 & \textbf{0.00E+00} & \textbf{0.00E+00} \\
\hline A11 & \textbf{0.00E+00} & \textbf{0.00E+00} & \textbf{0.00E+00} & \textbf{0.00E+00} & \textbf{0.00E+00} & \textbf{0.00E+00} & \textbf{0.00E+00} & \textbf{0.00E+00} & \textbf{0.00E+00} & \textbf{0.00E+00} & \textbf{0.00E+00} & \textbf{0.00E+00} & 1.17E-01 & \textbf{0.00E+00} & \textbf{0.00E+00} & \textbf{0.00E+00} & 8.55E-02 & \textbf{0.00E+00} & \textbf{0.00E+00} \\
\hline A12 & \textbf{0.00E+00} & 1.42E-14 & 1.73E+01 & \textbf{0.00E+00} & \textbf{0.00E+00} & 1.11E-14 & \textbf{0.00E+00} & 1.98E-09 & \textbf{0.00E+00} & \textbf{0.00E+00} & \textbf{0.00E+00} & \textbf{0.00E+00} & 2.02E+00 & \textbf{0.00E+00} & \textbf{0.00E+00} & \textbf{0.00E+00} & 2.91E+01 & \textbf{0.00E+00} & \textbf{0.00E+00} \\
\hline A13 & \textbf{0.00E+00} & \textbf{0.00E+00} & 2.38E-14 & \textbf{0.00E+00} & \textbf{0.00E+00} & \textbf{0.00E+00} & \textbf{0.00E+00} & \textbf{0.00E+00} & \textbf{0.00E+00} & \textbf{0.00E+00} & \textbf{0.00E+00} & \textbf{0.00E+00} & 1.92E-01 & \textbf{0.00E+00} & \textbf{0.00E+00} & \textbf{0.00E+00} & 4.12E-01 & \textbf{0.00E+00} & \textbf{0.00E+00} \\
\hline A14 & \textbf{0.00E+00} & \textbf{0.00E+00} & 2.38E-14 & \textbf{0.00E+00} & \textbf{0.00E+00} & \textbf{0.00E+00} & \textbf{0.00E+00} & \textbf{0.00E+00} & \textbf{0.00E+00} & \textbf{0.00E+00} & \textbf{0.00E+00} & \textbf{0.00E+00} & 1.92E-01 & \textbf{0.00E+00} & \textbf{0.00E+00} & \textbf{0.00E+00} & 4.12E-01 & \textbf{0.00E+00} & \textbf{0.00E+00} \\
\hline A15 & \textbf{0.00E+00} & \textbf{0.00E+00} & \textbf{0.00E+00} & \textbf{0.00E+00} & \textbf{0.00E+00} & \textbf{0.00E+00} & \textbf{0.00E+00} & \textbf{0.00E+00} & \textbf{0.00E+00} & \textbf{0.00E+00} & \textbf{0.00E+00} & \textbf{0.00E+00} & 3.95E-01 & \textbf{0.00E+00} & \textbf{0.00E+00} & \textbf{0.00E+00} & 9.71E+00 & \textbf{0.00E+00} & \textbf{0.00E+00} \\
\hline A16 & \textbf{0.00E+00} & \textbf{0.00E+00} & \textbf{0.00E+00} & \textbf{0.00E+00} & \textbf{0.00E+00} & \textbf{0.00E+00} & \textbf{0.00E+00} & \textbf{0.00E+00} & \textbf{0.00E+00} & \textbf{0.00E+00} & \textbf{0.00E+00} & \textbf{0.00E+00} & 4.05E+00 & \textbf{0.00E+00} & \textbf{0.00E+00} & \textbf{0.00E+00} & 4.19E+00 & \textbf{0.00E+00} & \textbf{0.00E+00} \\
\hline A17 & \textbf{0.00E+00} & \textbf{0.00E+00} & \textbf{0.00E+00} & \textbf{0.00E+00} & \textbf{0.00E+00} & \textbf{0.00E+00} & \textbf{0.00E+00} & \textbf{0.00E+00} & \textbf{0.00E+00} & \textbf{0.00E+00} & \textbf{0.00E+00} & \textbf{0.00E+00} & 3.99E+00 & \textbf{0.00E+00} & \textbf{0.00E+00} & \textbf{0.00E+00} & 7.04E-01 & \textbf{0.00E+00} & \textbf{0.00E+00} \\
\hline A18 & \textbf{0.00E+00} & \textbf{0.00E+00} & \textbf{0.00E+00} & \textbf{0.00E+00} & \textbf{0.00E+00} & \textbf{0.00E+00} & \textbf{0.00E+00} & \textbf{0.00E+00} & \textbf{0.00E+00} & \textbf{0.00E+00} & \textbf{0.00E+00} & \textbf{0.00E+00} & 3.52E-01 & \textbf{0.00E+00} & \textbf{0.00E+00} & \textbf{0.00E+00} & 6.19E-01 & \textbf{0.00E+00} & \textbf{0.00E+00} \\
\hline A19 & \textbf{0.00E+00} & 2.35E+01 & 9.07E+01 & 1.89E+01 & 1.45E-12 & 1.44E-12 & 1.19E-11 & 2.42E+01 & 1.15E+02 & \textbf{0.00E+00} & 1.19E+02 & 3.73E+01 & 1.20E+02 & 5.47E+01 & 2.86E-08 & 2.02E+02 & 2.32E+02 & 4.98E+01 & 5.50E-11 \\
\hline A20 & 1.27E-10 & 3.64E+01 & 4.08E+00 & 1.08E+02 & 8.32E-10 & 2.94E+00 & 2.55E+00 & 4.14E-06 & 3.28E+02 & 1.62E+01 & 3.47E+02 & 7.34E+01 & 9.12E+01 & 1.34E+02 & 1.76E+01 & 1.81E+02 & 3.26E+02 & 9.68E+01 & 2.11E+01 \\
\hline A21 & \textbf{0.00E+00} & 4.08E+01 & 6.79E+01 & 5.45E+02 & \textbf{0.00E+00} & 1.95E+01 & 1.17E+05 & 2.49E+01 & 3.94E+02 & 1.83E+01 & 3.86E+02 & 1.04E+02 & 1.12E+02 & 4.51E+02 & 1.22E+03 & 2.37E+02 & 3.89E+02 & 1.74E+02 & 1.56E+01 \\
\hline A22 & 1.39E-10 & 3.60E+01 & 9.27E+00 & 1.13E+02 & 7.58E-10 & 2.48E+00 & 1.35E+00 & 3.58E-06 & 3.28E+02 & 1.62E+01 & 3.47E+02 & 7.61E+01 & 9.44E+01 & 9.46E+01 & 1.82E+01 & 1.81E+02 & 5.25E+02 & 8.91E+01 & 1.90E+01 \\
\hline A23 & \textbf{0.00E+00} & 3.81E-12 & 2.67E+01 & \textbf{0.00E+00} & \textbf{0.00E+00} & \textbf{0.00E+00} & \textbf{0.00E+00} & 1.72E-01 & 7.05E-02 & \textbf{0.00E+00} & 1.07E-02 & \textbf{0.00E+00} & 1.83E+01 & 7.37E-08 & \textbf{0.00E+00} & \textbf{0.00E+00} & 2.26E+00 & \textbf{0.00E+00} & \textbf{0.00E+00} \\
\hline A24 & \textbf{0.00E+00} & 1.42E-14 & 5.35E+01 & \textbf{0.00E+00} & \textbf{0.00E+00} & \textbf{0.00E+00} & \textbf{0.00E+00} & 1.06E-01 & \textbf{0.00E+00} & \textbf{0.00E+00} & \textbf{0.00E+00} & \textbf{0.00E+00} & 8.55E+00 & \textbf{0.00E+00} & \textbf{0.00E+00} & \textbf{0.00E+00} & 5.12E+00 & \textbf{0.00E+00} & \textbf{0.00E+00} \\
\hline A25 & \textbf{0.00E+00} & 1.78E-12 & 1.88E+01 & \textbf{0.00E+00} & \textbf{0.00E+00} & 6.79E-14 & \textbf{0.00E+00} & 2.73E-05 & \textbf{0.00E+00} & \textbf{0.00E+00} & \textbf{0.00E+00} & \textbf{0.00E+00} & 3.41E-01 & \textbf{0.00E+00} & \textbf{0.00E+00} & \textbf{0.00E+00} & 2.87E+01 & \textbf{0.00E+00} & \textbf{0.00E+00} \\
\hline A26 & \textbf{0.00E+00} & 2.52E-12 & 1.72E+01 & \textbf{0.00E+00} & \textbf{0.00E+00} & 6.79E-14 & \textbf{0.00E+00} & 2.60E-05 & \textbf{0.00E+00} & \textbf{0.00E+00} & \textbf{0.00E+00} & \textbf{0.00E+00} & 3.36E-01 & \textbf{0.00E+00} & \textbf{0.00E+00} & \textbf{0.00E+00} & 1.91E+00 & \textbf{0.00E+00} & \textbf{0.00E+00} \\
\hline A28 & \textbf{0.00E+00} & 1.78E-12 & 6.27E+01 & \textbf{0.00E+00} & \textbf{0.00E+00} & 7.51E-14 & \textbf{0.00E+00} & 3.75E-05 & \textbf{0.00E+00} & \textbf{0.00E+00} & \textbf{0.00E+00} & \textbf{0.00E+00} & 7.51E-01 & \textbf{0.00E+00} & \textbf{0.00E+00} & \textbf{0.00E+00} & 3.94E+01 & \textbf{0.00E+00} & \textbf{0.00E+00} \\
\hline A29 & \textbf{0.00E+00} & 1.82E-12 & 1.72E+01 & \textbf{0.00E+00} & \textbf{0.00E+00} & 6.79E-14 & \textbf{0.00E+00} & 2.14E-05 & \textbf{0.00E+00} & \textbf{0.00E+00} & \textbf{0.00E+00} & \textbf{0.00E+00} & 1.06E+00 & \textbf{0.00E+00} & \textbf{0.00E+00} & \textbf{0.00E+00} & 3.26E+01 & \textbf{0.00E+00} & \textbf{0.00E+00} \\
\hline A30 & \textbf{0.00E+00} & 1.78E-12 & 1.88E+01 & \textbf{0.00E+00} & \textbf{0.00E+00} & 6.79E-14 & \textbf{0.00E+00} & 2.74E-05 & \textbf{0.00E+00} & \textbf{0.00E+00} & \textbf{0.00E+00} & \textbf{0.00E+00} & 3.41E-01 & \textbf{0.00E+00} & \textbf{0.00E+00} & \textbf{0.00E+00} & 2.87E+01 & \textbf{0.00E+00} & \textbf{0.00E+00} \\
\hline A31 & \textbf{0.00E+00} & 1.42E-14 & 5.35E+01 & \textbf{0.00E+00} & \textbf{0.00E+00} & \textbf{0.00E+00} & \textbf{0.00E+00} & 1.06E-01 & \textbf{0.00E+00} & \textbf{0.00E+00} & \textbf{0.00E+00} & \textbf{0.00E+00} & 8.55E+00 & \textbf{0.00E+00} & \textbf{0.00E+00} & \textbf{0.00E+00} & 5.12E+00 & \textbf{0.00E+00} & \textbf{0.00E+00} \\
\hline A32 & \textbf{0.00E+00} & 1.78E-12 & 1.88E+01 & \textbf{0.00E+00} & \textbf{0.00E+00} & 6.79E-14 & \textbf{0.00E+00} & 2.74E-05 & \textbf{0.00E+00} & \textbf{0.00E+00} & \textbf{0.00E+00} & \textbf{0.00E+00} & 3.41E-01 & \textbf{0.00E+00} & \textbf{0.00E+00} & \textbf{0.00E+00} & 2.87E+01 & \textbf{0.00E+00} & \textbf{0.00E+00} \\
\hline A33 & \textbf{0.00E+00} & 1.42E-14 & 5.35E+01 & \textbf{0.00E+00} & \textbf{0.00E+00} & \textbf{0.00E+00} & \textbf{0.00E+00} & 1.06E-01 & \textbf{0.00E+00} & \textbf{0.00E+00} & \textbf{0.00E+00} & \textbf{0.00E+00} & 8.55E+00 & \textbf{0.00E+00} & \textbf{0.00E+00} & \textbf{0.00E+00} & 5.12E+00 & \textbf{0.00E+00} & \textbf{0.00E+00} \\
\hline A34 & \textbf{0.00E+00} & 1.75E-12 & 3.07E+01 & \textbf{0.00E+00} & \textbf{0.00E+00} & 7.51E-14 & \textbf{0.00E+00} & 1.99E-05 & \textbf{0.00E+00} & \textbf{0.00E+00} & \textbf{0.00E+00} & \textbf{0.00E+00} & 2.06E+00 & \textbf{0.00E+00} & \textbf{0.00E+00} & \textbf{0.00E+00} & 4.48E+01 & \textbf{0.00E+00} & \textbf{0.00E+00} \\
\hline A35 & 8.97E+02 & 1.47E+01 & 4.53E+06 & 3.12E+02 & 8.89E+00 & 6.35E+00 & 2.44E+01 & 6.37E+03 & 2.00E+02 & 9.82E+01 & 1.93E+02 & 3.88E+02 & 9.83E+05 & 2.45E+02 & 3.54E+01 & 2.89E+02 & 6.61E+02 & 9.09E+01 & 5.76E+01 \\
\hline A36 & \textbf{0.00E+00} & 1.75E-12 & 1.88E+01 & \textbf{0.00E+00} & \textbf{0.00E+00} & 7.51E-14 & \textbf{0.00E+00} & 2.16E-05 & \textbf{0.00E+00} & \textbf{0.00E+00} & \textbf{0.00E+00} & \textbf{0.00E+00} & 2.06E+00 & \textbf{0.00E+00} & \textbf{0.00E+00} & \textbf{0.00E+00} & 4.48E+01 & \textbf{0.00E+00} & \textbf{0.00E+00} \\
\hline A37 & \textbf{0.00E+00} & 1.75E-12 & 1.88E+01 & \textbf{0.00E+00} & \textbf{0.00E+00} & 7.51E-14 & \textbf{0.00E+00} & 2.16E-05 & \textbf{0.00E+00} & \textbf{0.00E+00} & \textbf{0.00E+00} & \textbf{0.00E+00} & 2.06E+00 & \textbf{0.00E+00} & \textbf{0.00E+00} & \textbf{0.00E+00} & 4.48E+01 & \textbf{0.00E+00} & \textbf{0.00E+00} \\
\hline A38 & \textbf{0.00E+00} & 1.75E-12 & 1.88E+01 & \textbf{0.00E+00} & \textbf{0.00E+00} & 7.51E-14 & \textbf{0.00E+00} & 2.16E-05 & \textbf{0.00E+00} & \textbf{0.00E+00} & \textbf{0.00E+00} & \textbf{0.00E+00} & 2.06E+00 & \textbf{0.00E+00} & \textbf{0.00E+00} & \textbf{0.00E+00} & 4.48E+01 & \textbf{0.00E+00} & \textbf{0.00E+00} \\
\hline A39 & \textbf{0.00E+00} & 1.75E-12 & 1.88E+01 & \textbf{0.00E+00} & \textbf{0.00E+00} & 7.51E-14 & \textbf{0.00E+00} & 2.16E-05 & \textbf{0.00E+00} & \textbf{0.00E+00} & \textbf{0.00E+00} & \textbf{0.00E+00} & 2.06E+00 & \textbf{0.00E+00} & \textbf{0.00E+00} & \textbf{0.00E+00} & 4.48E+01 & \textbf{0.00E+00} & \textbf{0.00E+00} \\
\hline A40 & \textbf{0.00E+00} & \textbf{0.00E+00} & 1.08E-09 & \textbf{0.00E+00} & \textbf{0.00E+00} & \textbf{0.00E+00} & \textbf{0.00E+00} & \textbf{0.00E+00} & \textbf{0.00E+00} & \textbf{0.00E+00} & \textbf{0.00E+00} & \textbf{0.00E+00} & 2.54E-01 & \textbf{0.00E+00} & \textbf{0.00E+00} & \textbf{0.00E+00} & 4.50E+01 & \textbf{0.00E+00} & \textbf{0.00E+00} \\
\hline A41 & \textbf{0.00E+00} & 1.42E-14 & 2.46E+01 & \textbf{0.00E+00} & \textbf{0.00E+00} & \textbf{0.00E+00} & \textbf{0.00E+00} & 2.09E-01 & \textbf{0.00E+00} & \textbf{0.00E+00} & \textbf{0.00E+00} & \textbf{0.00E+00} & 8.52E+00 & \textbf{0.00E+00} & \textbf{0.00E+00} & \textbf{0.00E+00} & 3.21E+01 & \textbf{0.00E+00} & \textbf{0.00E+00} \\
\hline A42 & \textbf{0.00E+00} & 1.78E-12 & 1.88E+01 & \textbf{0.00E+00} & \textbf{0.00E+00} & 7.51E-14 & \textbf{0.00E+00} & 3.15E-05 & \textbf{0.00E+00} & \textbf{0.00E+00} & \textbf{0.00E+00} & \textbf{0.00E+00} & 5.60E-01 & \textbf{0.00E+00} & \textbf{0.00E+00} & \textbf{0.00E+00} & 1.97E+01 & \textbf{0.00E+00} & \textbf{0.00E+00} \\
\hline 
\end{tabular}
}
\end{table*}
\end{landscape}

\begin{figure}
\centering
\begin{subfigure}[b]{0.45\textwidth}
\includegraphics[scale=0.25]{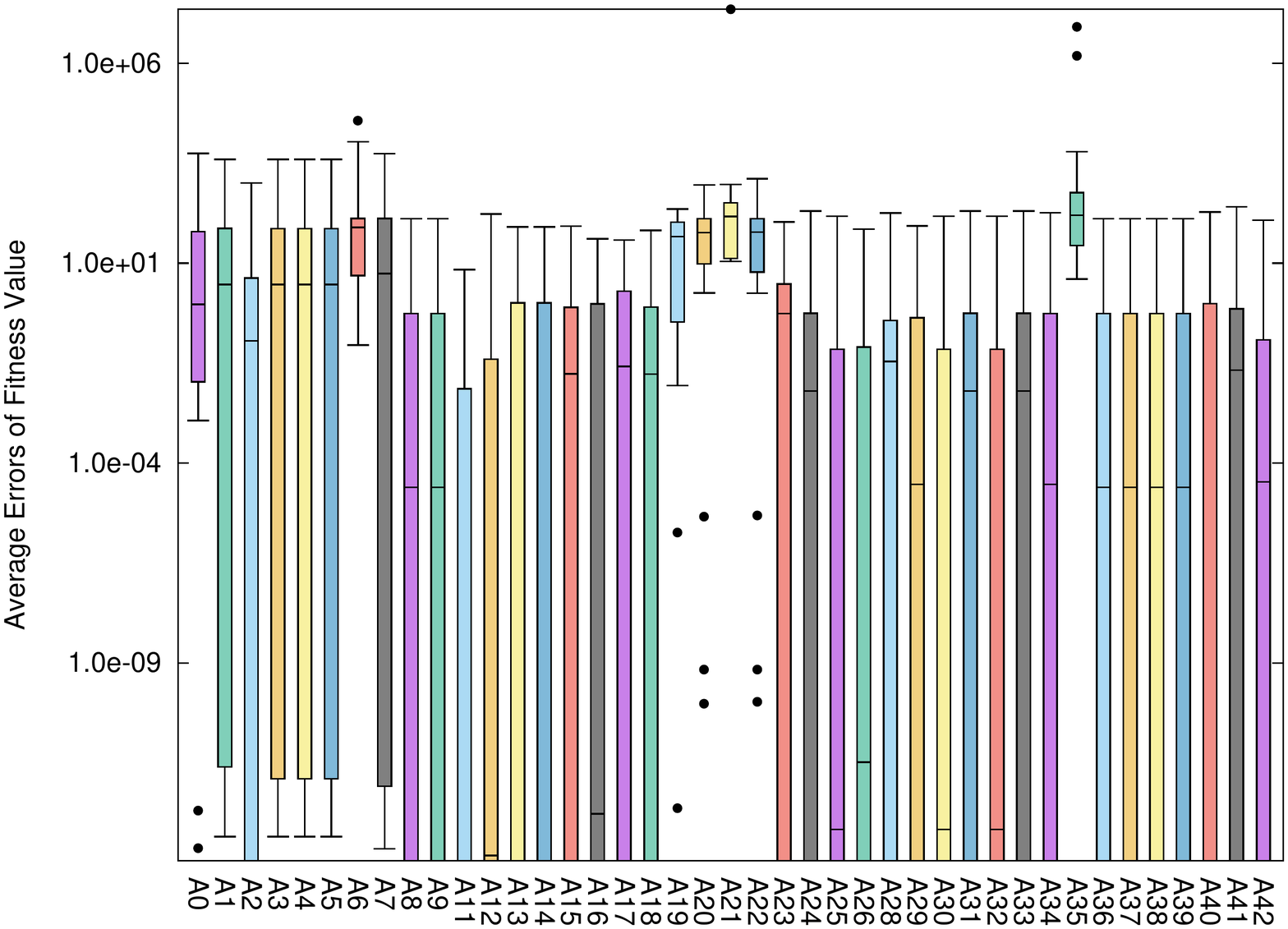}
\caption{Average error}
\label{nlopt_avg}
\end{subfigure}
\begin{subfigure}[b]{0.45\textwidth}
\includegraphics[scale=0.25]{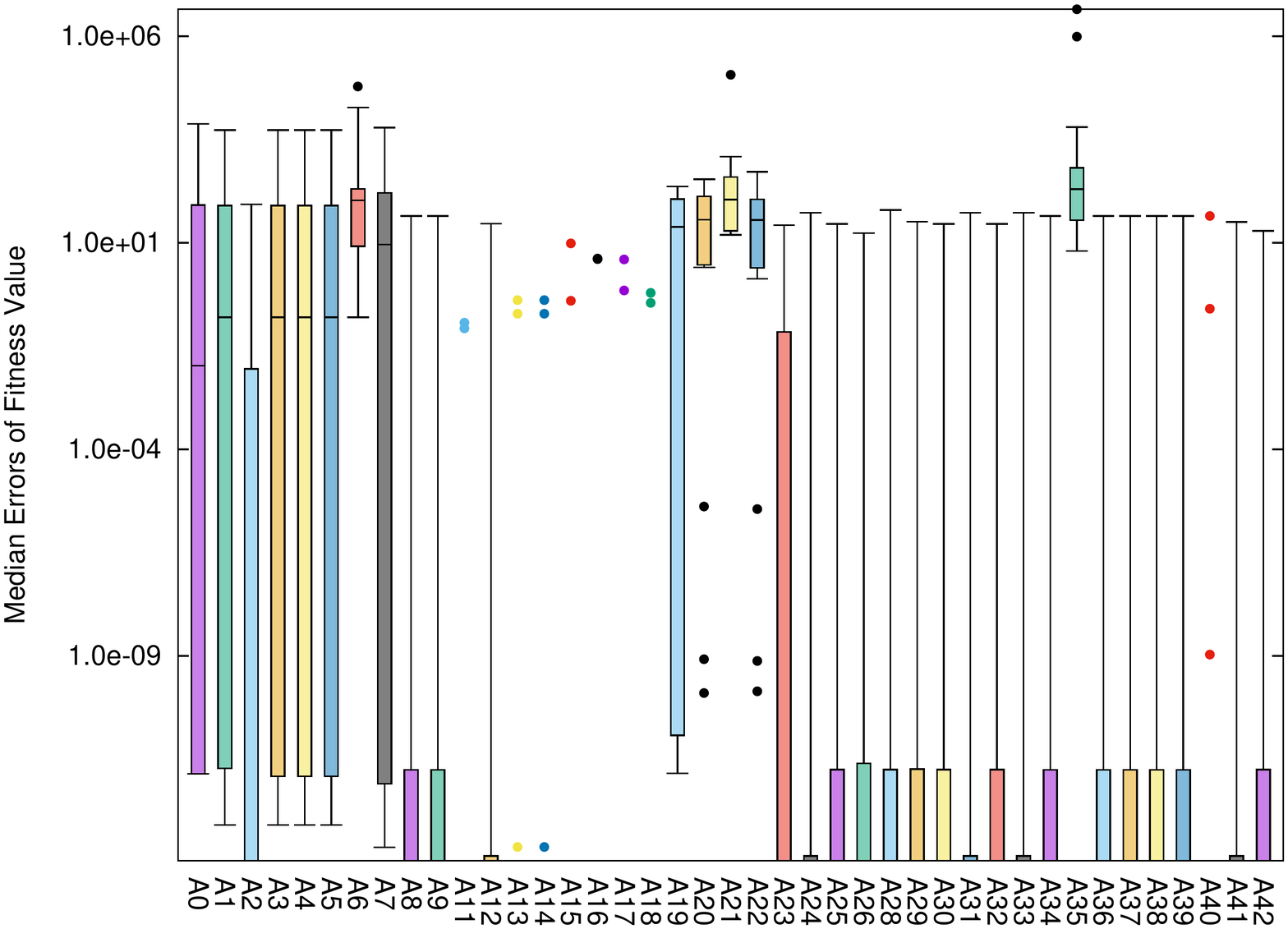}
\caption{Median error}
\label{nlopt_med}
\end{subfigure}
\caption{Box plots for average and median errors obtained on the SOCO benchmark functions.}
\label{soco_error}
\end{figure}

Further, the box plots for the average and median errors are shown in Figures \ref{nlopt_avg} and \ref{nlopt_med} respectively. The horizontal axis represents the algorithms from the NLopt library referred by the numeric identifiers as given in Table \ref{nlopt_algos}, and the vertical axis represents the mean and median errors in the two cases. It may be noted that all plots indicating mean/median errors use a logarithmic scale. One can infer from the plots that in terms of average error, the choice of algorithms A6, A19, A20, A21, A22 and A35 provides better performance, whereas in case of median error, the algorithms A11, A13, A14, A15, A16, A17, A18 and A40 demonstrate superior performance.

\begin{figure}
\centering
\begin{subfigure}[b]{0.45\textwidth}
\includegraphics[scale=0.25]{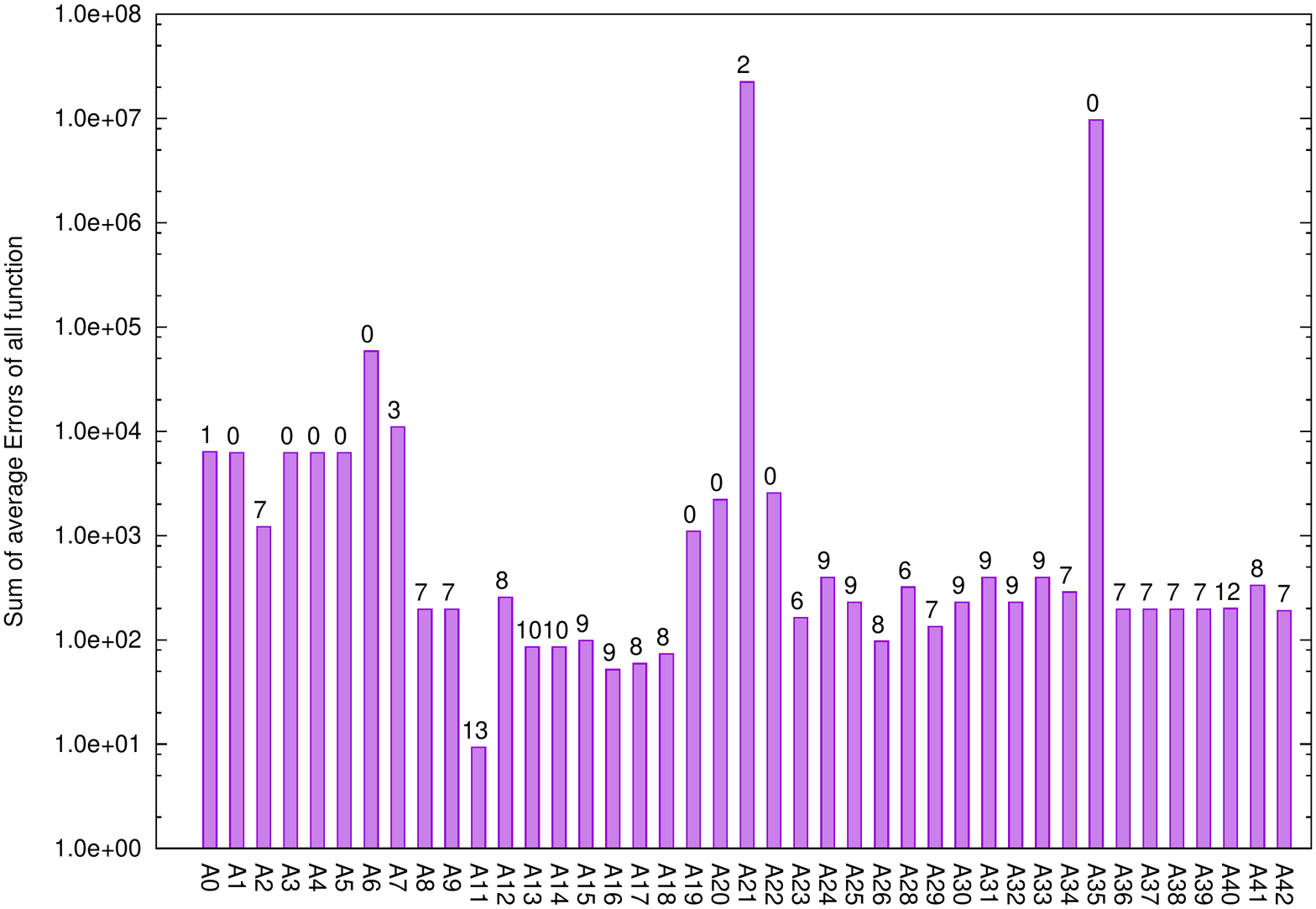}
\caption{Sum of average errors.}
\label{nlopt_avg_sum}
\end{subfigure}
\begin{subfigure}[b]{0.45\textwidth}
\includegraphics[scale=0.25]{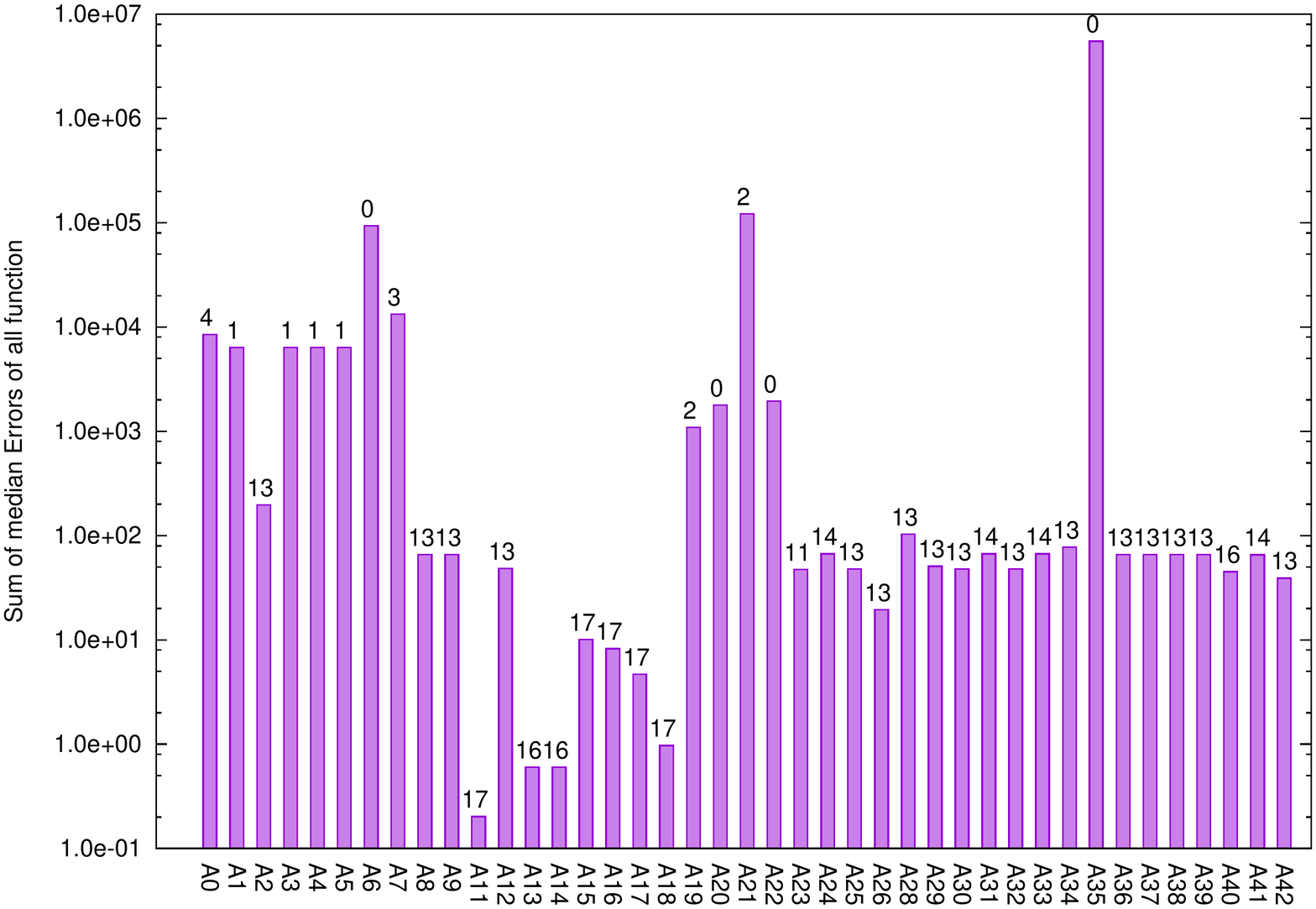}
\caption{Sum of median errors.}
\label{nlopt_med_sum}
\end{subfigure}
\caption{Sum of average and median errors on SOCO benchmarks for the NLopt algorithms. The number above the bar represents the number of times optima were found.}
\label{fig:soco_error_ranks}
\end{figure}

The sum of average and median errors are shown in Figures (\ref{nlopt_avg_sum}) and (\ref{nlopt_med_sum}) respectively. The horizontal axis indicates the algorithm from the NLopt library, while the vertical axis shows the sum of the corresponding average and median errors. The number at the top of each bar in the graph indicates the number of times the global minima were obtained on the SOCO benchmark functions. In terms of average error, algorithms A11 and A40 are able to find the global minima for 13 and 12 functions out of the 19 benchmarks, respectively. In case of median error, algorithms A11, A15, A16, A17 and A18 are able to find global minima for 17 out of the 19 benchmark functions. None of the reference algorithms have achieved zero median error on as many functions.

\begin{figure}
\centering
\includegraphics[scale=0.4]{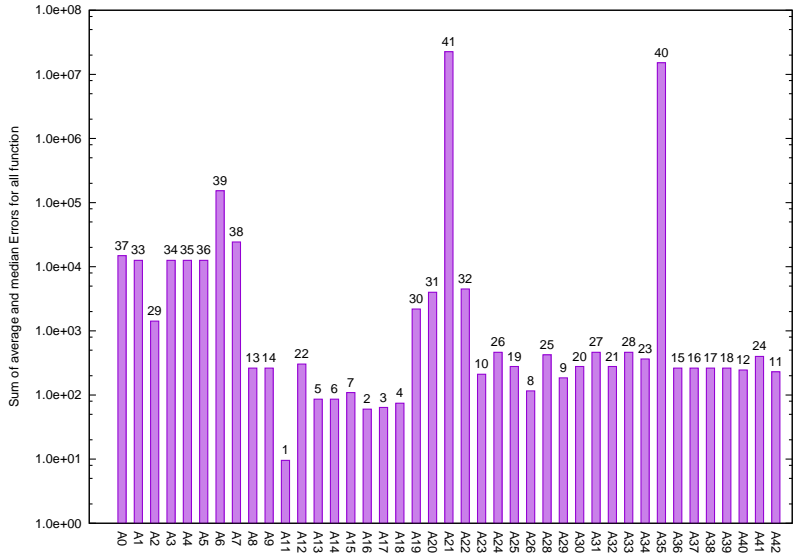}
\caption{Ranking of all algorithms based upon the sum of average and median errors on SOCO benchmarks}
\label{nlopt_rank}
\end{figure}

Figure \ref{nlopt_rank} aims to provide a ranking of the algorithms from the NLopt library on our approach. The ranking is done on the basis of the sum of average and median errors obtained on the benchmark functions, which is indicated on the vertical axis. The bars represent the value of sum of average and median errors for the algorithms on the horizontal axis. The rank of the algorithm is indicated above the bars. From our analysis, the top three algorithms for our approach are A11 (limited memory BFGS), A16 (Truncated Newton method with restart) and A17 (Preconditioned Truncated Newton), all of which belong to the class of gradient-based methods for local search. The worst performing algorithm from our analysis is A21 (MLSL-random) which belongs to the category of gradient-based global search algorithms of the NLopt library.

Finally, we also present a comparison of the performance of the best 3 NLopt algorithms (A11, A16 and A17) on the SOCO benchmark functions vis-\`a-vis reference algorithms available in literature. These include the Differential Evolution algorithm (DE) \cite{storn1997differential} and its variants, the co-variance matrix adaptation evolution strategy with increasing population size (G-CMA-ES) \cite{auger2005restart}, the real-coded CHC algorithm (CHC) \cite{eshelman1993chapter}, Shuffle Or Update Parallel Differential Evolution (SOUPDE) \cite{weber2011shuffle}, $DE-D^40+M^m$ \cite{garcia2011role}, Generalized Opposition-based Differential Evolution (GODE) \cite{wang2011enhanced}, Generalized Adaptive Differential Evolution (GADE) \cite{yang2011scalability}, jDElscop \cite{brest2011self}, Self-adaptive Differential Evolution with Multi-Trajectory Search (SaDE-MMTS) \cite{zhao2011self}, MOS-DE \cite{latorre2011mos}, MA-SSW-Chains \cite{molina2011memetic}, Restart Particle Swarm Optimization with Velocity Modulation (RPSO-VM) \cite{garcia2011restart}, Tuned IPSOLS \cite{de2011incremental} EVO-PROpt \cite{duarte2011path}, EM323 \cite{gardeux2011em323} and VXQR \cite{neumaier2011vxqr} among others.

\begin{figure*}[!htbp]
\centering
\begin{subfigure}[b]{0.45\textwidth}
 \includegraphics[width=\textwidth]{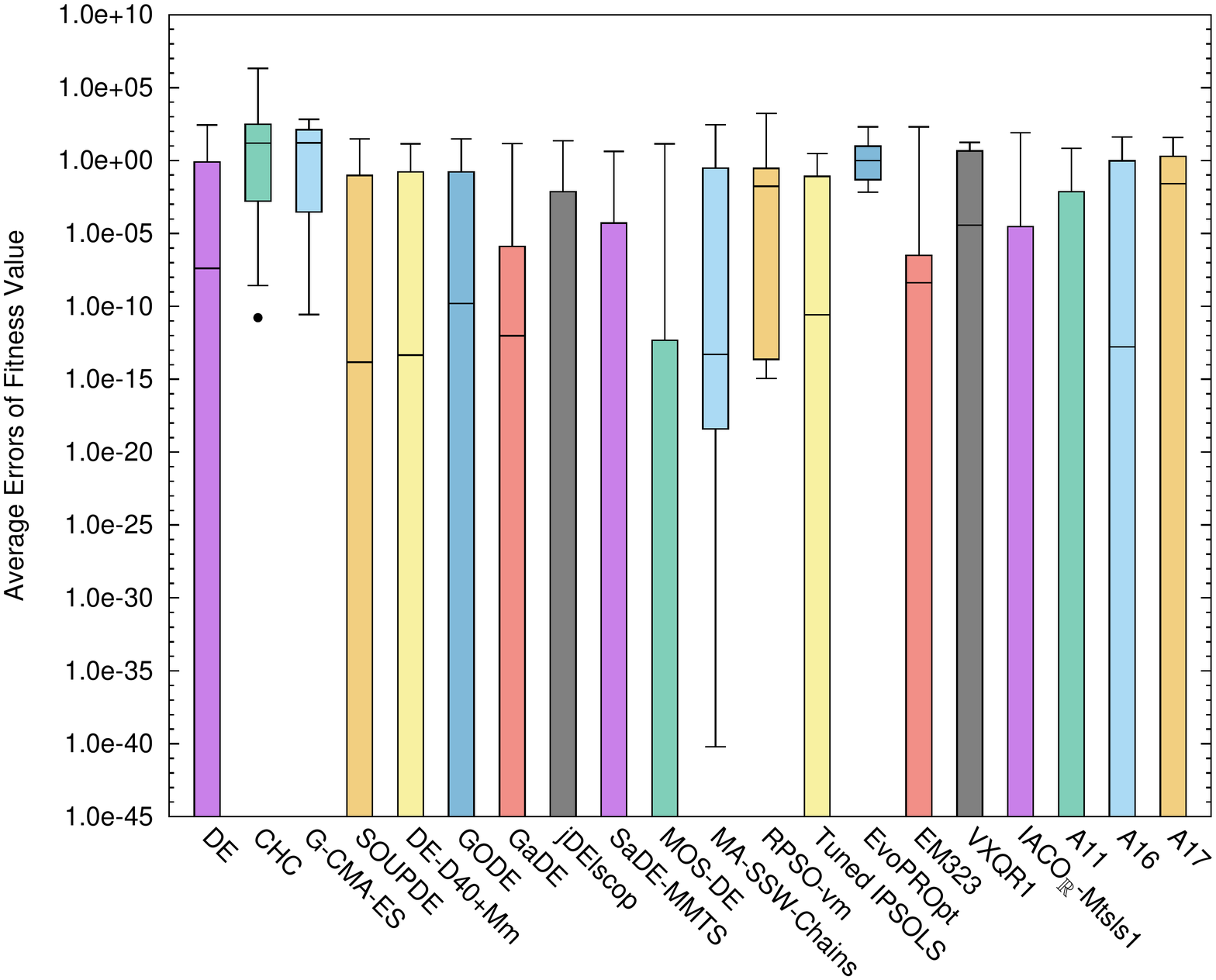}
 \caption{Average Error}
 \label{fig:avg_50_soco_comp}
\end{subfigure}
\begin{subfigure}[b]{0.45\textwidth}
 \includegraphics[width=\textwidth]{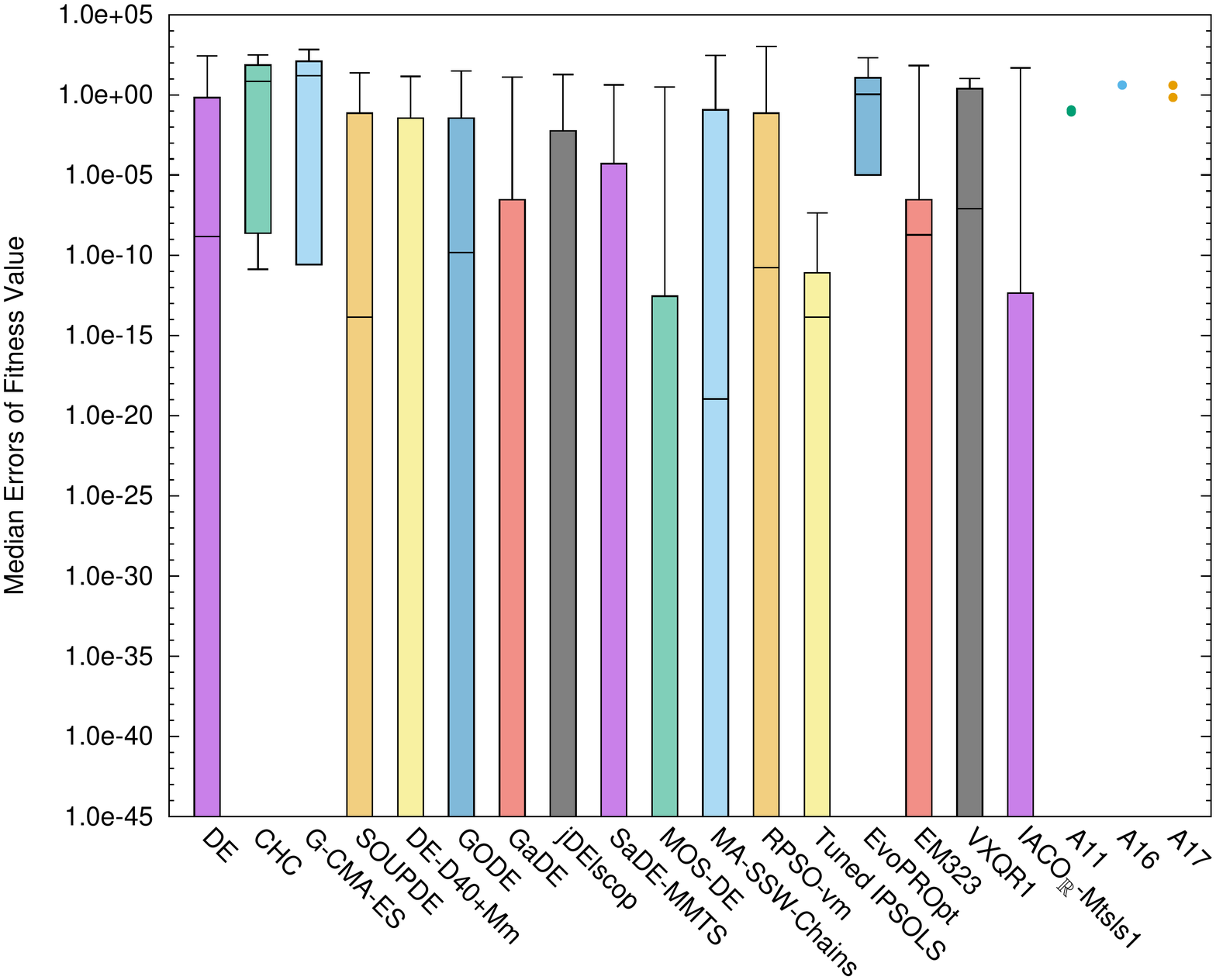}
 \caption{Median Error}
 \label{fig:med_50_soco_comp}
\end{subfigure}
\caption{Plots comparing average and median error of SOCO functions on reference algorithms and best 3 NLopt algorithms.}
\label{soco_compare_avg_med}
\end{figure*}

\begin{figure}[h]
\centering
\includegraphics[scale=0.3]{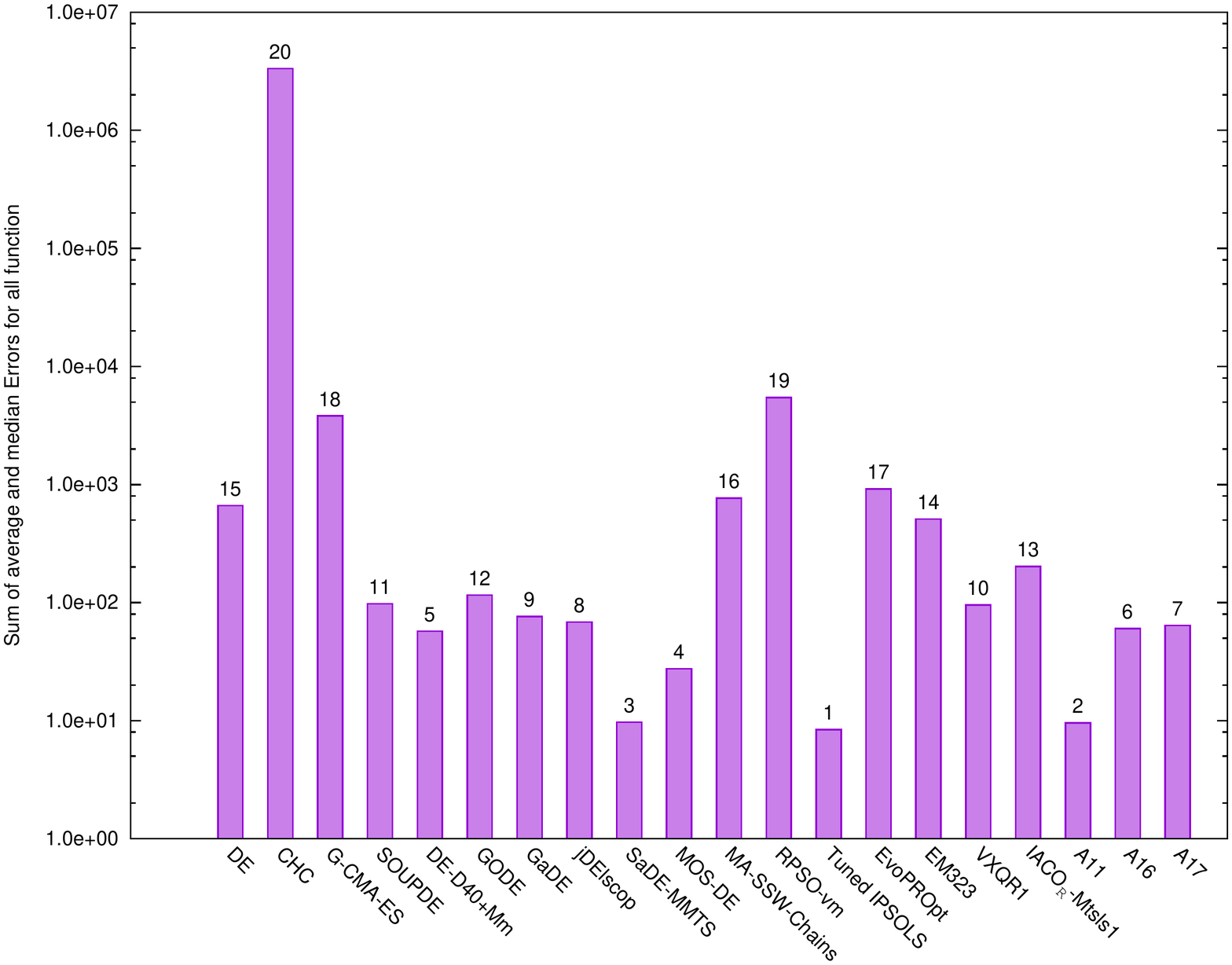}
\caption{Ranking of best 3 NLopt and reference algorithms on SOCO benchmarks.}
\label{nlopt_rank_best}
\end{figure}

The box plots of average and median error when comparing with $IACO_\mathbb{R}$-Mtsls1 are shown in Figure \ref{soco_compare_avg_med}. It can be inferred that the error obtained using NLopt algorithms are much lower than the reference algorithms. A ranking of the best 3 NLopt algorithms with the reference algorithms is shown in Figure \ref{nlopt_rank_best}. The best performing algorithm was found to be the Tuned-IPSOLS, whereas CHC performs the worst.

\subsection{Results on CEC 2014 benchmarks}

We now present the results on the CEC 2014 benchmarks \cite{liang2013problem}, where we have considered the 50-dimensional versions of the functions. The maximum number of function evaluations allowed was set to $10000 \times D$, where $D$ represents the number of dimensions in which the function is considered. The search range is $[-100,100]^D$, and uniform random initialization within the search space has been done. The algorithm was run $51$ times on each function; error values were defined as $f(x)-f(x^*)$, where $x$ is a candidate solution and $x^*$ is the optimal solution. Error values lower than $10^{-8}$ are approximated to 0. A summary of the benchmark functions is presented in Table \ref{tab:cec2014}.

\begin{table}[htbp]
  \centering
  \caption{CEC 2014 Benchmark Functions \cite{chenproblem}}
  \scalebox{0.9}{
    \begin{tabular}{|c|c|c|}
    \hline
    \textbf{Category} & \textbf{S. No. } & \multicolumn{1}{|c|}{\textbf{Function}} \\
    \hline
    \multirow{3}[0]{*}{Unimodal Functions} & 1     & Rotated High Conditioned Elliptic Function \\
          & 2     & Rotated Bent Cigar Function \\
          & 3     & Rotated Discus Function \\
          \hline
    \multirow{13}[0]{*}{Multimodal Functions} & 4     & Shifted and Rotated Rosenbrock’s Function \\
          & 5     & Shifted and Rotated Ackley’s Function \\
          & 6     & Shifted and Rotated Weierstrass Function \\
          & 7     & Shifted and Rotated Griewank’s Function \\
          & 8     & Shifted Rastrigin’s Function \\
          & 9     & Shifted and Rotated Rastrigin’s Function \\
          & 10    & Shifted Schwefel’s Function \\
          & 11    & Shifted and Rotated Schwefel’s Function \\
          & 12    & Shifted and Rotated Katsuura Function \\
          & 13    & Shifted and Rotated HappyCat Function \\
          & 14    & Shifted and Rotated HGBat Function \\
          & 15    & Shifted and Rotated Expanded Griewank’s plus Rosenbrock’s Function \\
          & 16    & Shifted and Rotated Expanded Scaffer’s F6 Function \\
          \hline
    \multirow{6}[0]{*}{Hybrid Function - 1} & 17    & Hybrid Function 1 (N=3) \\
          & 18    & Hybrid Function 2 (N=3) \\
          & 19    & Hybrid Function 3 (N=4) \\
          & 20    & Hybrid Function 4 (N=4) \\
          & 21    & Hybrid Function 5 (N=5) \\
          & 22    & Hybrid Function 6 (N=5) \\
          \hline
    \multirow{8}[0]{*}{Composition Functions} & 23    & Composition Function 1 (N=5) \\
          & 24    & Composition Function 2 (N=3) \\
          & 25    & Composition Function 3 (N=3) \\
          & 26    & Composition Function 4 (N=5) \\
          & 27    & Composition Function 5 (N=5) \\
          & 28    & Composition Function 6 (N=5) \\
          & 29    & Composition Function 7 (N=3) \\
          & 30    & Composition Function 8 (N=3) \\
    \hline
    \end{tabular}%
    }
  \label{tab:cec2014}%
\end{table}%

\begin{figure*}[!htbp]
\centering
\begin{subfigure}[b]{0.45\textwidth}
 \includegraphics[width=\textwidth]{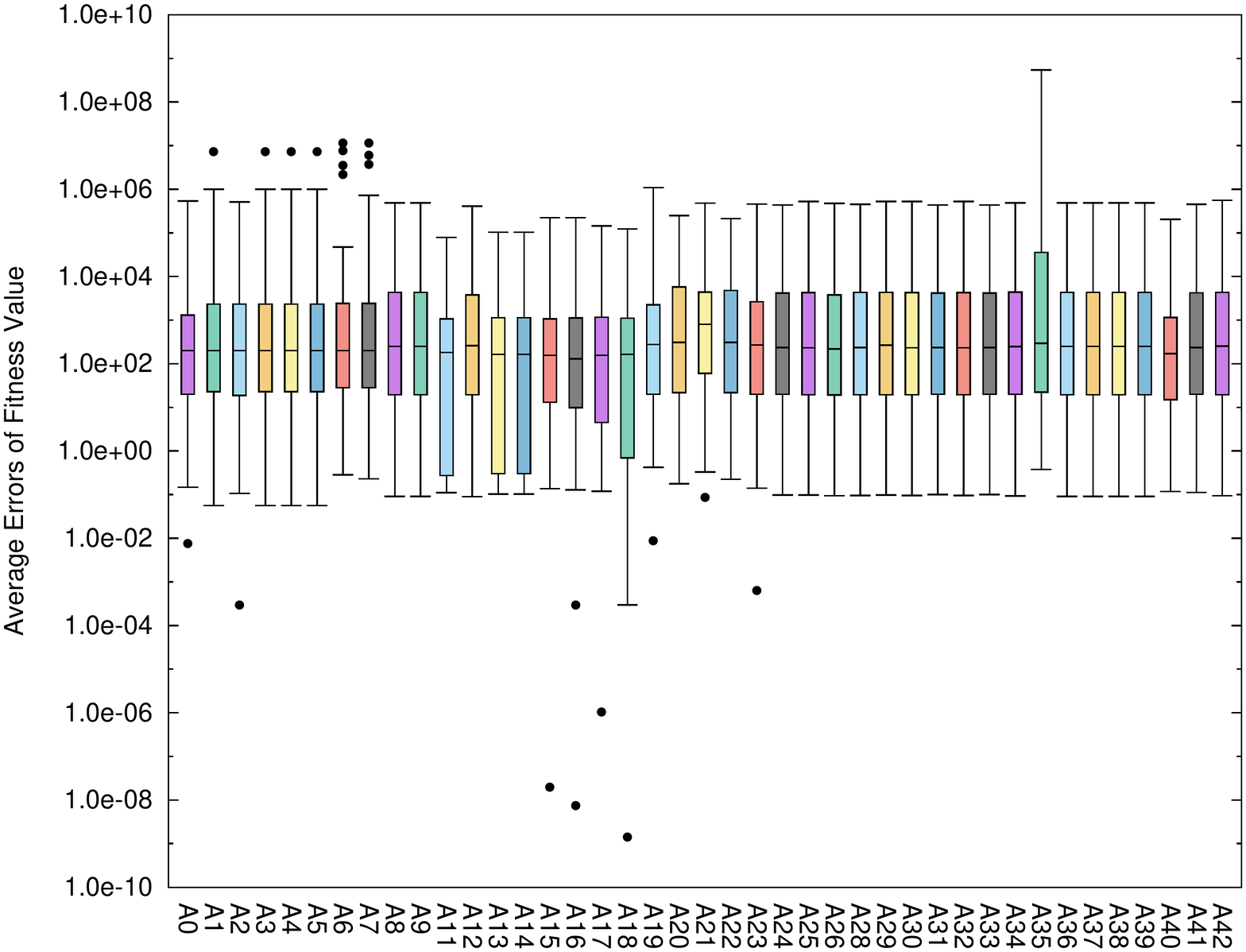}
 \caption{Average Error}
 \label{avg_cec2014}
\end{subfigure}
\begin{subfigure}[b]{0.45\textwidth}
 \includegraphics[width=\textwidth]{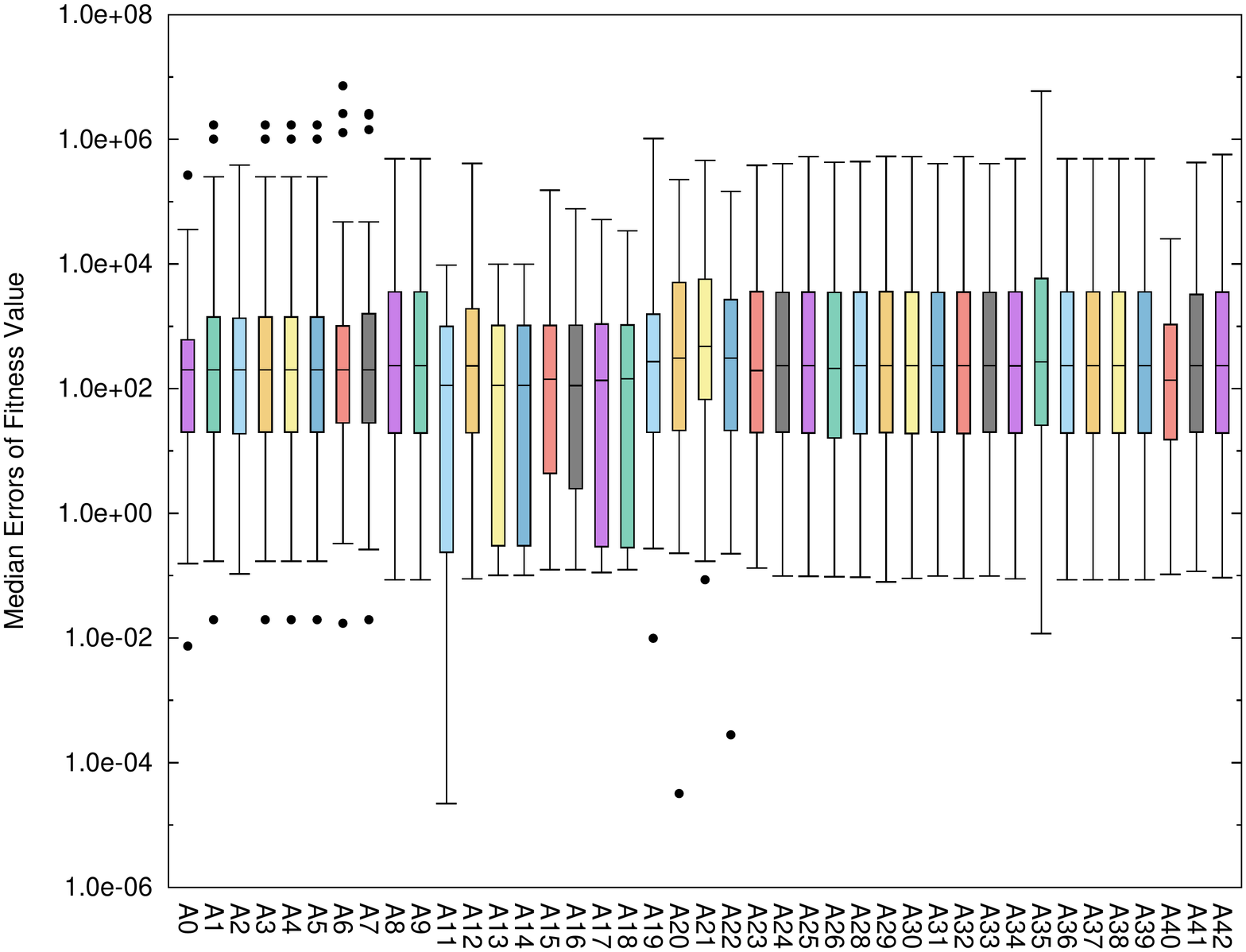}
 \caption{Median Error}
 \label{median_cec2014}
\end{subfigure}
\begin{subfigure}[b]{0.45\textwidth}
 \includegraphics[width=\textwidth]{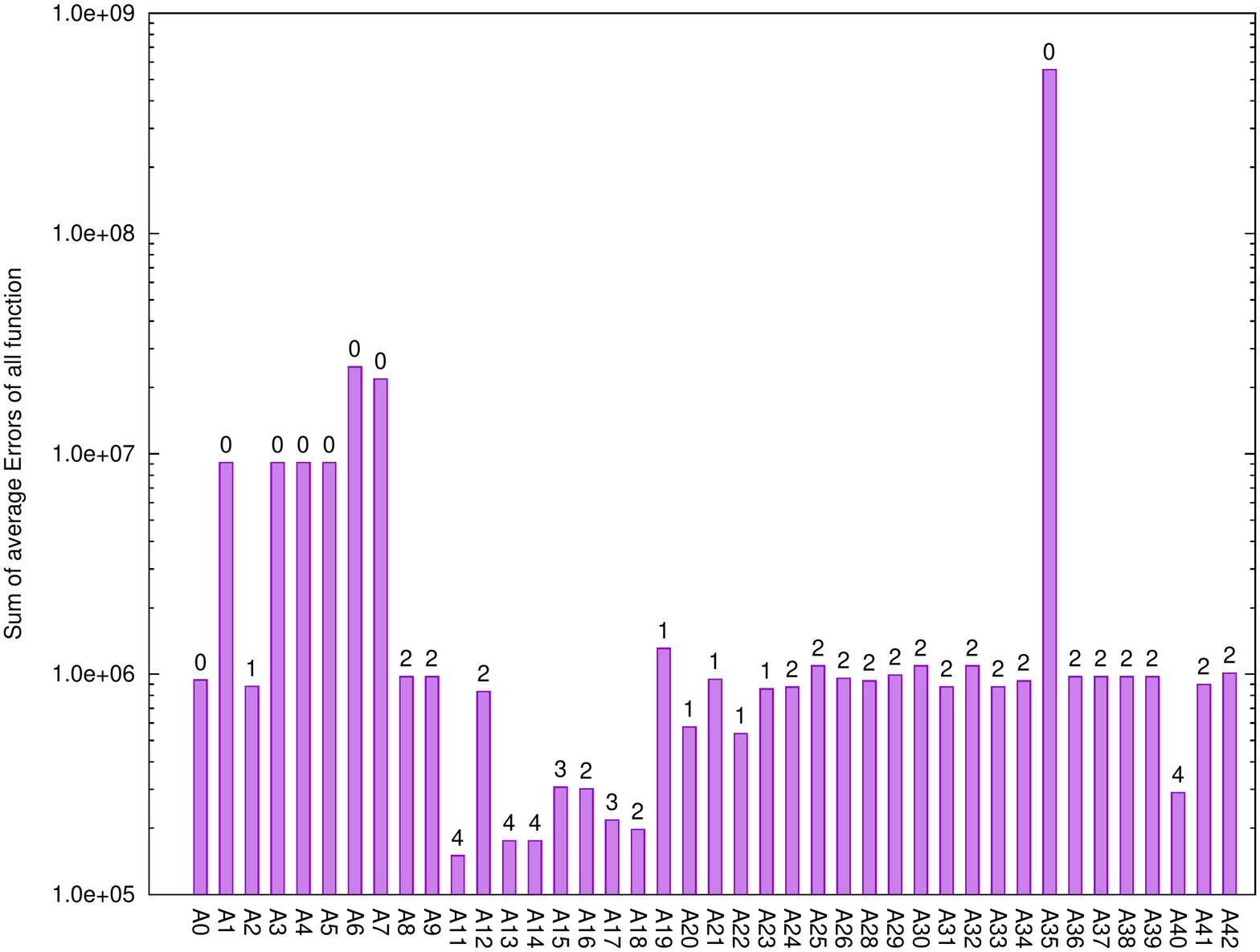}
 \caption{Average Error Ranking}
 \label{avg_cec2014_rank}
\end{subfigure}
\begin{subfigure}[b]{0.45\textwidth}
 \includegraphics[width=\textwidth]{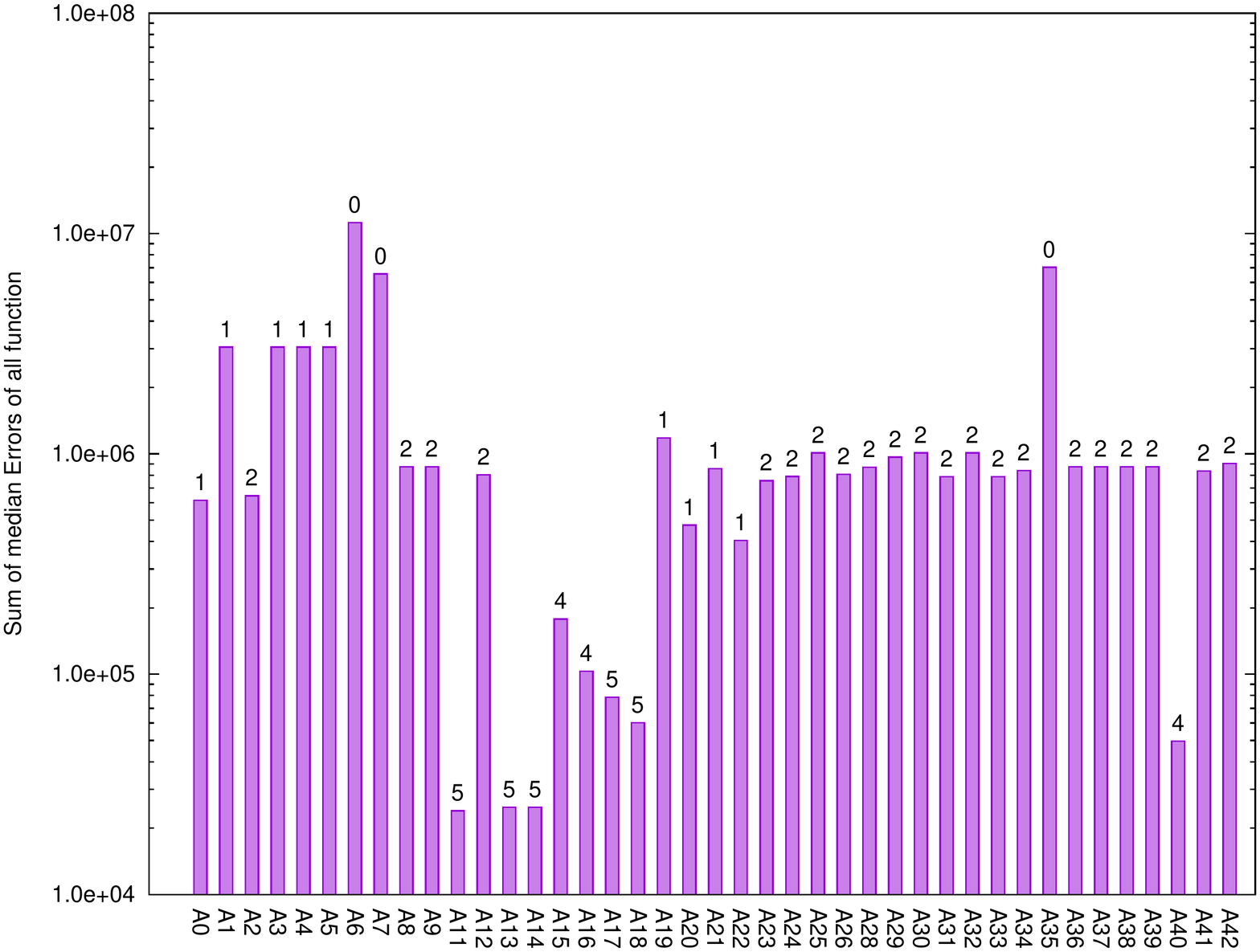}
 \caption{Median Error Ranking}
 \label{median_cec2014_rank}
\end{subfigure}
\caption{Plots showing average and median error of NLopt algorithms with their ranking on CEC 2014 benchmarks.}
\label{fig:cec2014_errors}
\end{figure*}

The average and median errors of the NLopt algorithms on the CEC benchmarks are shown in Figure \ref{fig:cec2014_errors}. The box plots showing average error are shown in Figure \ref{avg_cec2014}, while median error is shown in Figure \ref{median_cec2014}. It may be noted here that the plots are for the 50-dimensional versions of the functions. Further, the sum of average and median errors are shown in Figures \ref{avg_cec2014_rank} and \ref{median_cec2014_rank} respectively.

\begin{figure*}[!htbp]
\centering
 \includegraphics[scale=0.25]{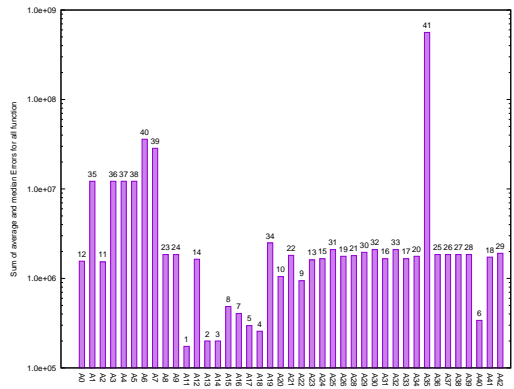}
 \caption{Ranking of the NLopt algorithms on the CEC 2014 benchmarks.}
\label{fig:algo_ranking_cec2014}
\end{figure*}

The ranking of the algorithms on the CEC 2014 benchmarks is shown in Figure \ref{fig:algo_ranking_cec2014}. One can observe that A11 (limited memory BFGS) performs the best while A35 (ISRES evolutionary constrained optimization) performs the worst.
 
We also provide a comparison of the top 3 NLopt algorithms (A11, A13 and A14) on the CEC 2014 benchmarks with other reference algorithms. These include the United Multi-Operator Evolutionary Algorithms (UMOEA) \cite{elsayed2014testing}, Success-History based Adaptive Differential Evolution using linear population size reduction (L-SHADE) \cite{tanabe2014improving}, Differential Evolution with Replacement Strategy (RSDE) \cite{xu2014differential}, Memetic Differential Evolution Based on Fitness Euclidean-Distance Ratio (FERDE) \cite{qu2014memetic}, Partial Opposition-Based Adaptive Differential Evolution (POBL-ADE) \cite{hu2014partial}, Mean-Variance Mapping Optimization (MVMO) \cite{erlich2014evaluating}, rmalschcma \cite{molina2014influence}, Opt Bees \cite{maia2014real}, Fireworks Algorithm with Differential Mutation (FWA-DE) \cite{yu2014fireworks}, Non-uniform Real-coded Genetic Algorithm (NRGA) \cite{yashesh2014non}, b3e3pbest \cite{bujok2014differential} and DE\_b6e6rl \cite{polakova2014controlled}.
 
\begin{figure*}[!htbp]
\centering
\begin{subfigure}[b]{0.45\textwidth}
 \includegraphics[width=\textwidth]{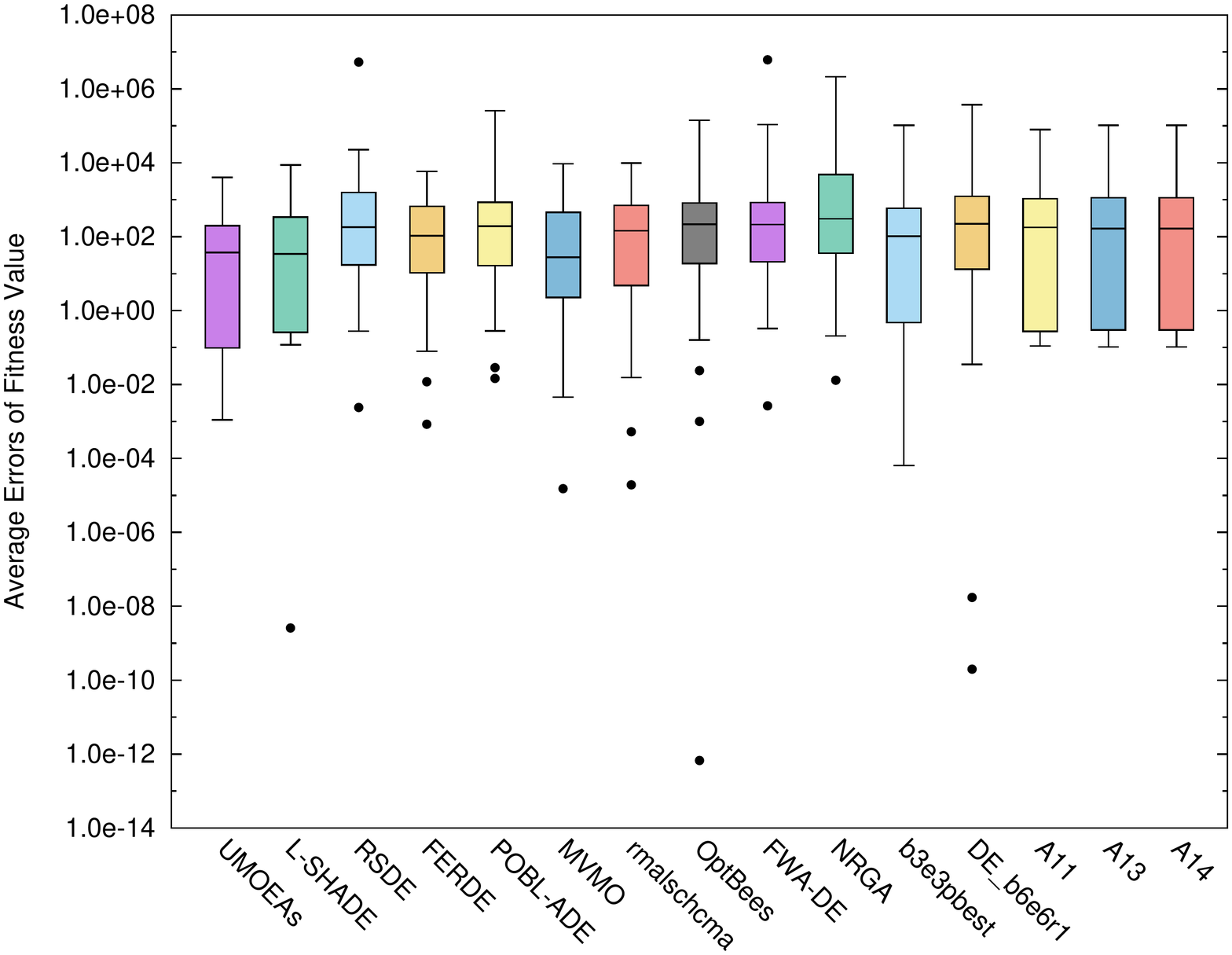}
 \caption{Average Error}
 \label{avg_cec2014_comp}
\end{subfigure}
\begin{subfigure}[b]{0.45\textwidth}
 \includegraphics[width=\textwidth]{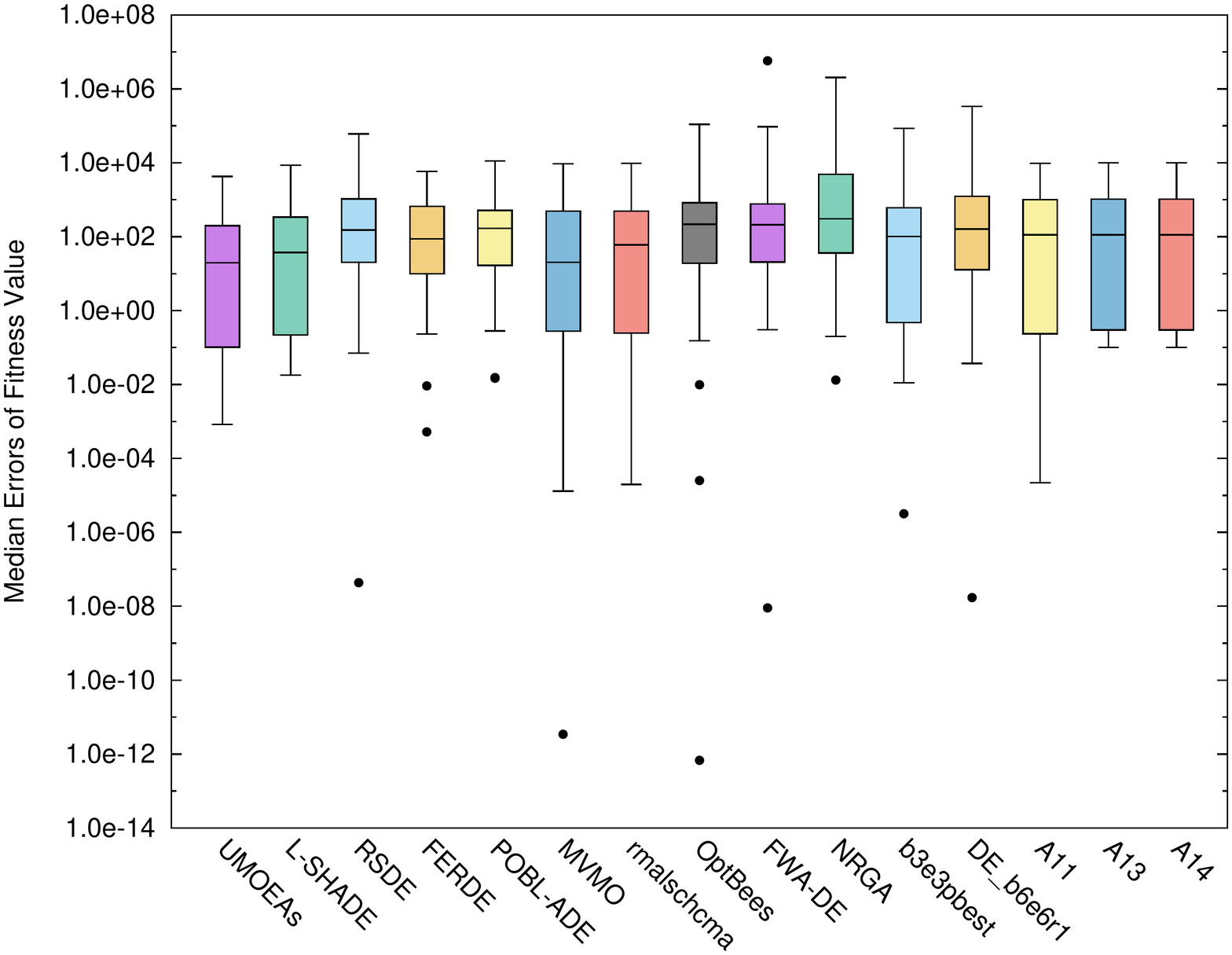}
 \caption{Median Error}
 \label{median_cec2014_comp}
\end{subfigure}
\caption{Plots showing average and median error of best 3 NLopt and reference algorithms on CEC 2014 benchmarks.}
\label{fig:cec2014_errors_comp}
\end{figure*}

The box plots for average and median errors are shown in Figure \ref{fig:cec2014_errors_comp}, specifically average error in Figure (\ref{avg_cec2014_comp}) and median error in Figure (\ref{median_cec2014_comp}) respectively. The range of average errors of the reference algorithms are relatively lower than the top 3 NLopt algorithms except on UMOEA, and also median error except MVMO and   rmalschcma.

\begin{figure*}[!htbp]
\centering
 \includegraphics[scale=0.3]{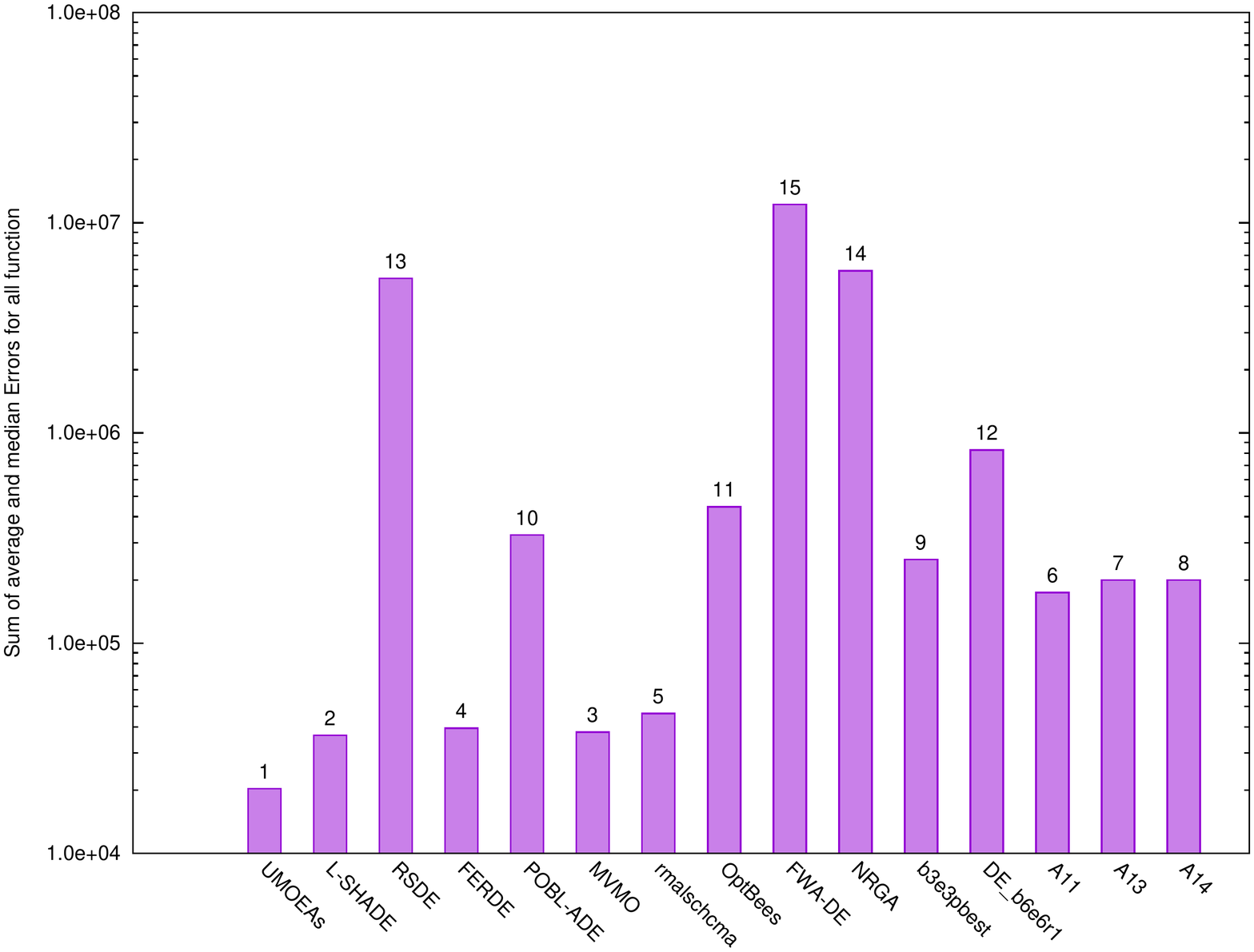}
 \caption{Ranking of the top 3 NLopt and reference algorithms on the CEC 2014 benchmarks.}
\label{fig:algo_ranking_cec2014_comp}
\end{figure*}

A relative ranking of these algorithms is shown in Figure \ref{fig:algo_ranking_cec2014_comp}. One can observe that the best performing algorithm is UMOEA, while the worst performing algorithm is FWA-DE.

\subsection{Comparison of Hybrid approach with standalone algorithms used for Local Search}

We also provide a comparison of the standalone performance of the algorithms in Table \ref{nlopt_pval_tab}. The comparison is provided in terms of Wilcoxon Signed-Rank test \cite{wilcoxon1945individual}, which is a measure of the extent of statistical deviations in the results obtained using a particular approach. A p-value less than $0.05$ indicates that the results of our approach have a significant statistical difference with the results obtained using the algorithms being compared, whereas p-values greater than $0.05$ indicate non-significant statistical difference. The columns $(Wp)$ and $(Wn)$ denote the sum of signed ranks. The column $(N)$ indicates the number of instances for which there is a difference in the result between two algorithms.

It can be observed from Table \ref{nlopt_pval_tab} that on the SOCO benchmark functions, the NLopt algorithms give significant statistical difference on \textbf{all but one} (A34) in terms of average error, and on \textbf{all but two} (A6 and A7) in terms of median error. On the CEC 2014 benchmarks, there is significant statistical difference on all but \textbf{8} and \textbf{3} out of \textbf{42} algorithms in terms of average and median error respectively. 

\begin{table*}
\centering
\caption{Wilcoxon signed rank test between hybrid and standalone approach}
\label{nlopt_pval_tab}
\scalebox{0.68}{
\begin{tabular} {|c|c|c|c|c|c|c|c|c|c|c|c|c|c|c|c|c|}
\hline & \multicolumn{8}{|c|} {\textbf{SOCO}} & \multicolumn{8}{|c|} {\textbf{CEC2014}} \\
\hline  & \multicolumn{4}{|c|} {\textit{Average}}  & \multicolumn{4}{|c|} {\textit{Median}}  & \multicolumn{4}{|c|} {\textit{Average}}  & \multicolumn{4}{|c|} {\textit{Median}}  \\
\hline  & \textit{Wp} & \textit{Wn} & \textit{n} & \textit{p} & \textit{Wp} & \textit{Wn} & \textit{n} & \textit{p} & \textit{Wp} & \textit{Wn} & \textit{n} & \textit{p} & \textit{Wp} & \textit{Wn} & \textit{n} & \textit{p} \\
\hline A0 & 7 & 183 & 19 & \textbf{3.98E-04} & 0 & 78 & 12 & \textbf{4.88E-04} & 26 & 250 & 23 & \textbf{6.58E-04} & 0 & 190 & 19 & \textbf{1.32E-04} \\
\hline A1 & 4 & 186 & 19 & \textbf{2.50E-04} & 0 & 91 & 13 & \textbf{2.44E-04} & 32 & 268 & 24 & \textbf{7.48E-04} & 0 & 153 & 17 & \textbf{2.93E-04} \\
\hline A2 & 0 & 190 & 19 & \textbf{1.32E-04} & 0 & 190 & 19 & \textbf{1.32E-04} & 26 & 274 & 24 & \textbf{3.96E-04} & 0 & 231 & 21 & \textbf{5.96E-05} \\
\hline A3 & 3 & 187 & 19 & \textbf{2.14E-04} & 0 & 91 & 13 & \textbf{2.44E-04} & 32 & 268 & 24 & \textbf{7.48E-04} & 0 & 153 & 17 & \textbf{2.93E-04} \\
\hline A4 & 3 & 187 & 19 & \textbf{2.14E-04} & 0 & 91 & 13 & \textbf{2.44E-04} & 32 & 268 & 24 & \textbf{7.48E-04} & 0 & 153 & 17 & \textbf{2.93E-04} \\
\hline A5 & 3 & 187 & 19 & \textbf{2.14E-04} & 0 & 91 & 13 & \textbf{2.44E-04} & 32 & 268 & 24 & \textbf{7.48E-04} & 0 & 153 & 17 & \textbf{2.93E-04} \\
\hline A6 & 0 & 190 & 19 & \textbf{1.32E-04} & 0 & 0 & 0 & 1.00E+00 & 33 & 243 & 23 & \textbf{1.41E-03} & 0 & 105 & 14 & \textbf{1.22E-04} \\
\hline A7 & 2 & 134 & 16 & \textbf{6.43E-04} & 0 & 0 & 0 & 1.00E+00 & 33 & 243 & 23 & \textbf{1.41E-03} & 0 & 91 & 13 & \textbf{2.44E-04} \\
\hline A8 & 0 & 190 & 19 & \textbf{1.32E-04} & 0 & 190 & 19 & \textbf{1.32E-04} & 61 & 404 & 30 & \textbf{4.20E-04} & 62 & 403 & 30 & \textbf{4.53E-04} \\
\hline A9 & 0 & 190 & 19 & \textbf{1.32E-04} & 0 & 190 & 19 & \textbf{1.32E-04} & 61 & 404 & 30 & \textbf{4.20E-04} & 62 & 403 & 30 & \textbf{4.53E-04} \\
\hline A11 & 1 & 135 & 16 & \textbf{5.31E-04} & 0 & 91 & 13 & \textbf{2.44E-04} & 109 & 216 & 25 & 1.50E-01 & 68 & 208 & 23 & \textbf{3.33E-02} \\
\hline A12 & 1 & 152 & 17 & \textbf{3.52E-04} & 1 & 135 & 16 & \textbf{5.31E-04} & 71 & 307 & 27 & \textbf{4.58E-03} & 72 & 306 & 27 & \textbf{4.94E-03} \\
\hline A13 & 0 & 153 & 17 & \textbf{2.93E-04} & 2 & 134 & 16 & \textbf{6.43E-04} & 109 & 242 & 26 & 9.12E-02 & 29 & 247 & 23 & \textbf{9.16E-04} \\
\hline A14 & 0 & 153 & 17 & \textbf{2.93E-04} & 2 & 134 & 16 & \textbf{6.43E-04} & 109 & 242 & 26 & 9.12E-02 & 29 & 247 & 23 & \textbf{9.16E-04} \\
\hline A15 & 0 & 136 & 16 & \textbf{4.38E-04} & 0 & 120 & 15 & \textbf{6.10E-05} & 93 & 313 & 28 & \textbf{1.23E-02} & 32 & 374 & 28 & \textbf{9.86E-05} \\
\hline A16 & 0 & 136 & 16 & \textbf{4.38E-04} & 0 & 120 & 15 & \textbf{6.10E-05} & 91 & 315 & 28 & \textbf{1.08E-02} & 34 & 344 & 27 & \textbf{1.96E-04} \\
\hline A17 & 3 & 150 & 17 & \textbf{5.03E-04} & 0 & 120 & 15 & \textbf{6.10E-05} & 93 & 285 & 27 & \textbf{2.11E-02} & 36 & 315 & 26 & \textbf{3.96E-04} \\
\hline A18 & 1 & 152 & 17 & \textbf{3.52E-04} & 0 & 105 & 14 & \textbf{1.22E-04} & 92 & 286 & 27 & \textbf{1.98E-02} & 29 & 271 & 24 & \textbf{5.46E-04} \\
\hline A19 & 15 & 175 & 19 & \textbf{1.28E-03} & 16 & 174 & 19 & \textbf{1.48E-03} & 147 & 288 & 29 & 1.27E-01 & 98 & 253 & 26 & 5.05E-02 \\
\hline A20 & 17 & 173 & 19 & \textbf{1.70E-03} & 7 & 183 & 19 & \textbf{3.98E-04} & 71 & 254 & 25 & \textbf{1.38E-02} & 62 & 238 & 24 & \textbf{1.19E-02} \\
\hline A21 & 0 & 153 & 17 & \textbf{2.93E-04} & 0 & 153 & 17 & \textbf{2.92E-04} & 53 & 325 & 27 & \textbf{1.08E-03} & 31 & 320 & 26 & \textbf{2.42E-04} \\
\hline A22 & 18 & 172 & 19 & \textbf{1.94E-03} & 13 & 177 & 19 & \textbf{9.67E-04} & 95 & 230 & 25 & 6.73E-02 & 78 & 247 & 25 & \textbf{2.30E-02} \\
\hline A23 & 0 & 153 & 17 & \textbf{2.93E-04} & 2 & 151 & 17 & \textbf{4.21E-04} & 66 & 340 & 28 & \textbf{1.81E-03} & 57 & 294 & 26 & \textbf{2.62E-03} \\
\hline A24 & 0 & 153 & 17 & \textbf{2.93E-04} & 0 & 153 & 17 & \textbf{2.93E-04} & 52 & 354 & 28 & \textbf{5.85E-04} & 44 & 307 & 26 & \textbf{8.38E-04} \\
\hline A25 & 0 & 190 & 19 & \textbf{1.32E-04} & 0 & 190 & 19 & \textbf{1.32E-04} & 99 & 336 & 29 & \textbf{1.04E-02} & 99 & 336 & 29 & \textbf{1.04E-02} \\
\hline A26 & 0 & 190 & 19 & \textbf{1.32E-04} & 0 & 190 & 19 & \textbf{1.32E-04} & 240 & 195 & 29 & 6.27E-01 & 229 & 236 & 30 & 9.34E-01 \\
\hline A28 & 0 & 190 & 19 & \textbf{1.32E-04} & 0 & 190 & 19 & \textbf{1.32E-04} & 83 & 382 & 30 & \textbf{2.11E-03} & 86 & 379 & 30 & \textbf{2.58E-03} \\
\hline A29 & 0 & 190 & 19 & \textbf{1.32E-04} & 0 & 190 & 19 & \textbf{1.32E-04} & 56 & 409 & 30 & \textbf{2.83E-04} & 88 & 377 & 30 & \textbf{2.96E-03} \\
\hline A30 & 0 & 190 & 19 & \textbf{1.32E-04} & 0 & 190 & 19 & \textbf{1.32E-04} & 99 & 336 & 29 & \textbf{1.04E-02} & 99 & 336 & 29 & \textbf{1.04E-02} \\
\hline A31 & 0 & 153 & 17 & \textbf{2.93E-04} & 0 & 153 & 17 & \textbf{2.93E-04} & 38 & 340 & 27 & \textbf{2.86E-04} & 33 & 318 & 26 & \textbf{2.96E-04} \\
\hline A32 & 0 & 190 & 19 & \textbf{1.32E-04} & 0 & 190 & 19 & \textbf{1.32E-04} & 99 & 336 & 29 & \textbf{1.04E-02} & 99 & 336 & 29 & \textbf{1.04E-02} \\
\hline A33 & 0 & 153 & 17 & \textbf{2.93E-04} & 0 & 153 & 17 & \textbf{2.93E-04} & 38 & 340 & 27 & \textbf{2.86E-04} & 33 & 318 & 26 & \textbf{2.96E-04} \\
\hline A34 & 38 & 115 & 17 & 6.84E-02 & 0 & 153 & 17 & \textbf{2.93E-04} & 253 & 182 & 29 & 4.43E-01 & 229 & 149 & 27 & 3.37E-01 \\
\hline A35 & 0 & 190 & 19 & \textbf{1.32E-04} & 9 & 181 & 19 & \textbf{5.39E-04} & 2 & 433 & 29 & \textbf{3.17E-06} & 0 & 435 & 29 & \textbf{2.56E-06} \\
\hline A36 & 0 & 190 & 19 & \textbf{1.32E-04} & 0 & 190 & 19 & \textbf{1.32E-04} & 61 & 404 & 30 & \textbf{4.20E-04} & 62 & 403 & 30 & \textbf{4.53E-04} \\
\hline A37 & 0 & 190 & 19 & \textbf{1.32E-04} & 0 & 190 & 19 & \textbf{1.32E-04} & 61 & 404 & 30 & \textbf{4.20E-04} & 62 & 403 & 30 & \textbf{4.53E-04} \\
\hline A38 & 0 & 190 & 19 & \textbf{1.32E-04} & 0 & 190 & 19 & \textbf{1.32E-04} & 61 & 404 & 30 & \textbf{4.20E-04} & 62 & 403 & 30 & \textbf{4.53E-04} \\
\hline A39 & 0 & 190 & 19 & \textbf{1.32E-04} & 0 & 190 & 19 & \textbf{1.32E-04} & 61 & 404 & 30 & \textbf{4.20E-04} & 62 & 403 & 30 & \textbf{4.53E-04} \\
\hline A40 & 0 & 136 & 16 & \textbf{4.38E-04} & 0 & 136 & 16 & \textbf{4.38E-04} & 99 & 226 & 25 & 8.75E-02 & 62 & 214 & 23 & \textbf{2.08E-02} \\
\hline A41 & 0 & 153 & 17 & \textbf{2.93E-04} & 0 & 153 & 17 & \textbf{2.93E-04} & 71 & 335 & 28 & \textbf{2.65E-03} & 38 & 313 & 26 & \textbf{4.79E-04} \\
\hline A42 & 0 & 190 & 19 & \textbf{1.32E-04} & 0 & 190 & 19 & \textbf{1.32E-04} & 37 & 428 & 30 & \textbf{5.79E-05} & 37 & 428 & 30 & \textbf{5.79E-05} \\
\hline
\end{tabular}
}
\end{table*}

\section{Conclusions }
\label{sec:conclusion}
This paper presented an exhaustive analysis of using optimization algorithms from the NLopt library in combination with the Mtsls1 algorithm within the $IACO_\mathbb{R}$-Mtsls1 framework for continuous global optimization. The results on SOCO and CEC 2014 benchmark functions present a ready reference on the performance of these approaches and would be of help to a researcher in deciding on a choice among these algorithms. The nature of functions on which these algorithms perform better can also be inferred from the results. A relative ranking of these approaches has also been provided based on the total error obtained in using them, which would provide a measure of the versatility of the algorithms. 

The results of our analysis have been summarized in Table \ref{final_res}. On both the benchmark function sets, the hybrid $IACO_\mathbb{R}$-Mtsls1 with gradient-based local search performs better (A11, A16, A17 for SOCO and A11, A13, A14 for CEC). The best results are obtained using hybridization with BFGS on both benchmarks. On SOCO benchmarks, the hybrid approach outperforms the original $IACO_\mathbb{R}$-Mtsls1, as it has been able to achieve zero median error on \textbf{17} out of \textbf{19} functions, which has not been achieved by any other algorithm considered. We believe that the analysis presented in this paper would be of use to the research community at large.

\begin{table}[H]
  \centering
  \caption{Summary of algorithm rankings.}
  \scalebox{0.7}{
    \begin{tabular}{|c|c||c|c|}
    \hline
    \textbf{Function} & \textbf{Ranking} & \textbf{NLopt} & \textbf{State-of-Art} \\
    \hline
    \textbf{SOCO} & Best  & A11   & Tuned IPSOLS \\
    \textbf{} & Worst & A21   & CHC \\
    \hline
    \textbf{CEC 2014} & Best  & A11   & UMOEA \\
    \textbf{} & Worst & A35   & FWA-DE \\
    \hline
    \multicolumn{4}{|l|}{\textit{A11: Limited Memory BFGS }} \\
    \multicolumn{4}{|l|}{\textit{A21: Multi-Level Single Linkage, random}} \\
    \multicolumn{4}{|l|}{\textit{A35: ISRES Evolutionary Constrained Optimization}} \\
    \multicolumn{4}{|l|}{\textit{Tuned IPSOLS: Incremental Particle Swarm for Large Scale Optimization}} \\
    \multicolumn{4}{|l|}{\textit{UMOEA: united Multi Operator Evolutionary Algorithms}} \\
    \multicolumn{4}{|l|}{\textit{FWA-DE: FireWorks Algorithm with Differential Evolution}} \\
    \hline
    \end{tabular}%
    }
  \label{final_res}%
\end{table}%

\bibliographystyle{plain}
\bibliography{ref}

\end{document}